%% file: main.tex
\documentclass[Afour,sageh,times]{sagej}

\usepackage{moreverb,url}

\usepackage[colorlinks,bookmarksopen,bookmarksnumbered,citecolor=red,urlcolor=red]{hyperref}
\usepackage{amsmath}
\usepackage{cleveref}
\crefname{figure}{Fig.}{Figs.}
\Crefname{figure}{Fig.}{Figs.}
\crefname{table}{Table}{Tables}
\Crefname{table}{Table}{Tables}

\usepackage{algorithm}
\usepackage{algorithmic}

\crefname{section}{Section}{Sections}
\Crefname{section}{Section}{Sections}
\usepackage{mathtools}
\usepackage{amssymb}
\usepackage[]{graphicx}
\usepackage[]{caption}
\usepackage[]{subcaption}

\usepackage[]{amsfonts}

\newcommand\BibTeX{{\rmfamily B\kern-.05em \textsc{i\kern-.025em b}\kern-.08em
T\kern-.1667em\lower.7ex\hbox{E}\kern-.125emX}}

\def\volumeyear{2026}
\begin{document}

\runninghead{Belmonte-Baeza et al.}

\title{Path Planning and Reinforcement Learning-Driven Control of On-Orbit Free-Flying Multi-Arm Robots}

\author{\'Alvaro Belmonte-Baeza\affilnum{1} and Jos\'e Luis Ram\'on\affilnum{1} and Leonard Felicetti \affilnum{2} and Miguel Cazorla \affilnum{1} and Jorge Pomares \affilnum{1}}

\affiliation{\affilnum{1}University of Alicante, Spain\\
\affilnum{2}Cranfield University, UK}

\corrauth{\'Alvaro Belmonte-Baeza, University of Alicante,
Ctra. San Vicente S/N,
San Vicente del Raspeig,
03690, Spain}

\email{alvaro.belmonte@ua.es}

\begin{abstract}

This paper presents a hybrid approach that integrates trajectory optimization (TO) and reinforcement learning (RL) for motion planning and control of free-flying multi-arm robots in on-orbit servicing scenarios. The proposed system integrates TO for generating feasible, efficient paths while accounting for dynamic and kinematic constraints, and RL for adaptive trajectory tracking under uncertainties. The multi-arm robot design, equipped with thrusters for precise body control, enables redundancy and stability in complex space operations. TO optimizes arm motions and thruster forces, reducing reliance on the arms for stabilization and enhancing maneuverability. RL further refines this by leveraging model-free control to adapt to dynamic interactions and disturbances. The experimental results validated through comprehensive simulations demonstrate the effectiveness and robustness of the proposed hybrid approach. Two case studies are explored: surface motion with initial contact and a free-floating scenario requiring surface approximation. In both cases, the hybrid method outperforms traditional strategies. In particular, the thrusters notably enhance motion smoothness, safety, and operational efficiency. The RL policy effectively tracks TO-generated trajectories, handling high-dimensional action spaces and dynamic mismatches. This integration of TO and RL combines the strengths of precise, task-specific planning with robust adaptability, ensuring high performance in the uncertain and dynamic conditions characteristic of space environments. By addressing challenges such as motion coupling, environmental disturbances, and dynamic control requirements, this framework establishes a strong foundation for advancing the autonomy and effectiveness of space robotic systems.

\end{abstract}


\maketitle

{\let\thefootnote\relax\footnotetext{\textbf{Preprint Note:} Accepted for publication in the \textit{International Journal of Robotics Research} (IJRR). Please cite the published version: Belmonte-Baeza \textit{et al.}, ``Path Planning and Reinforcement Learning-Driven Control of On-Orbit Free-Flying Multi-Arm Robots,'' \textit{International Journal of Robotics Research}, 2026. DOI: \textit{Will be added when available.}}}

\input{sections/introduction}
\input{sections/system_architecture_and_dynamics}

\input{sections/trajectory_optimization}

\input{sections/rl_framework}
\input{sections/results}

\input{sections/conclusions}

\subsection*{\normalsize\sagesf\bfseries Author contributions}

\subsection*{\normalsize\sagesf\bfseries Statements and Declarations}

\begin{dci}
    The authors declare no potential conflicts of interest with respect to the research, authorship, and/or publication of this article
\end{dci}

\begin{funding}
This research received funding from the European Union's European Innovation Council (EIC) Pathfinder programme under Grant Agreement No. 101223360 for the DEXTER project, by the project CIAICO/2022/077 (Programa AICO 2023, Conselleria de Innovación, Universidades, Ciencia y Sociedad Digital de la Generalitat Valenciana, Spain), and by the project PID2024-160373OB-C22 funded by MICIU /AEI /10.13039/501100011033 / FEDER, UE. Álvaro Belmonte-Baeza was supported by the Spanish Ministry of Universities under grant FPU21/02586. 
\end{funding}

\bibliographystyle{SageH}
\bibliography{references}

\clearpage
\section*{Appendix}
\subsection*{A.1 Robot arm dynamic and kinematic parameters}
\begin{table}[h]
\small\sf\centering
\caption{\bf Main dynamic parameters of the robot arms. Mass is in kg, length in meters, and all inertia terms are expressed in $\boldsymbol{kg \cdot m^2}$}
\label{tab:robot_params}
\begin{tabular}{c|c|c|c|c}
\toprule
\bfseries Link  & \bfseries \textbf{Mass} & \bfseries \textbf{Length} & \bfseries \textbf{Ixx / Iyy} & \bfseries \textbf{Izz} \\
\midrule
\textbf{Base} & 4.0 & 0.090 & 0.0044 & 0.0072  \\
\hline
\textbf{Shoulder} & 3.7 & 0.162 & 0.0103 & 0.0067 \\
\hline
\textbf{Upper Arm} & 8.393 & 0.425 & 0.2269 & 0.0151 \\
\hline
\textbf{Forearm} & 2.275 & 0.392 & 0.0494 & 0.0041 \\
\hline
\textbf{Wrist 1} & 1.219 & 0.133 & 0.1112 & 0.2194 \\
\hline
\textbf{Wrist 2} & 1.219 & 0.133 & 0.1112 & 0.2194 \\
\hline
\textbf{Wrist 3} & 0.188 & 0.046 & 0.0171 & 0.0338 \\
\bottomrule
\end{tabular}
\end{table}

\subsection*{A.2 Reward weights}
\begin{table}[h]
\small\sf\centering
\caption{\bf{Reward weights and regularization terms for training our RL policy}}
    \begin{tabular}{c|c}
    \toprule
    \bfseries \textbf{Reward term weight} & \bfseries \textbf{Value} \\
    \midrule
        $w_{body}$   & $20.0$ \\
        $w_{ee}$   & $15.0$ \\
        $\epsilon$   & $10^{-5}$ \\
        $w_{pow}$   & $-2.5\times{10^{-2}}$ \\
        $w_{acc}$   & $-1\times{10^{-6}}$ \\
        $w_{bacc}$   & $-1\times{10^{-2}}$ \\
        $w_{act}$   & $-1\times{10^{-2}}$ \\
        $w_{th}$   & $-1\times{10^{-2}}$ \\
        $w_{c}$   & $-1.0$ \\
        $w_{bc}$   & $-200.0$ \\
    \bottomrule
    \end{tabular}
    \label{tab:rlweights}
\end{table}

\subsection*{A.3 Domain Randomization and Noise Injection}
\begin{table}[h]
\small\sf\centering
\caption{\bf{Domain randomization values. All values represent the randomization added to the default value of the parameter (e.g. A random initialization of $-0.1m$ kg means that the mass is $90\%$ of the default mass).}}
    \begin{tabular}{c|c}
    \toprule
    \bfseries \textbf{Parameter} & \bfseries \textbf{Value} \\
    \midrule
        Initial body position & $\mathcal{U}(-0.2, 0.2)m$ \\
        Initial joint positions & $\mathcal{U}(-0.1, 0.1) rad$ \\
        System mass & $\mathcal{U}(-0.1m, 0.1m) kg$ \\
    \bottomrule
    \end{tabular}
    \label{tab:domain_randomization}
\end{table}

\begin{table}[h]
\small\sf\centering
\caption{\bf{Observation noise injection. All values represent the nosie added to the default value of the parameter}}
    \begin{tabular}{c|c}
    \toprule
    \bfseries \textbf{Observation} & \bfseries \textbf{Noise} \\
    \midrule
        $\boldsymbol{d}_{b}$ & $\mathcal{U}(-0.025, 0.025)m$ \\
        $\boldsymbol{\phi}_{b}$ & $\mathcal{U}(-0.02, 0.02)rad$ \\
        $\boldsymbol{v}_b$ & $\mathcal{U}(-0.05, 0.05) m/s$ \\
        $\boldsymbol{\omega}_b$ & $\mathcal{U}(-0.05, 0.05) rad/s$ \\
        $\boldsymbol{q}$ & $\mathcal{U}(-0.02, 0.02) rad$ \\
        $\boldsymbol{\dot{q}}$ & $\mathcal{U}(-0.1, 0.1) rad/s$ \\
    \bottomrule
    \end{tabular}
    \label{tab:noise}
\end{table}

\subsection*{A.4 RL Algorithm Hyperparameters}
\begin{table}[h]
\small\sf\centering
\caption{\bf{PPO Hyperparamenters}}
    \begin{tabular}{c|c}
    \toprule
    \bfseries \textbf{Parameter} & \bfseries \textbf{Value} \\
    \midrule
        Batch size & 196608 (8192x24) \\
        Mini-batch size & 49152 (8192x6) \\
        Number of epochs & 8 \\
        Clip range & 0.2 \\
        Entropy coefficient & 0.01 \\
        Discount factor $\gamma$ & 0.99 \\
        GAE discount factor & 0.95 \\
        Desired KL-Divergence & 0.01 \\
        Learning rate & 0.001 (adaptive) \\
    \bottomrule
    \end{tabular}
    \label{tab:ppo_params}
\end{table}

\end{document}

%% file: sections/introduction.tex
\section{Introduction}\label{sec:intro}

Space exploration has witnessed remarkable advances in the past years, leading to increasingly complex mission objectives such as satellite servicing, debris removal, and the assembly of large structures like space stations and telescopes \citep{Ma2023OOS}. These missions require sophisticated robotic systems capable of intricate manipulation tasks in the harsh and unpredictable environment of space. The environment is inherently non-deterministic and characterized by microgravity, orbital perturbations, and various disturbances that can significantly affect robotic operations. In addition, communication delays, limited computational resources, and uncertainties in the operational environment further complicate the path planning and control of robotic systems. There is thus a critical need for robots that can adapt to these uncertainties and maintain high performance in unpredictable workspaces while being highly autonomous.

Redundancy through multiple degrees of freedom allows space robots to navigate complex environments and overcome unforeseen obstacles. This redundancy is particularly beneficial for tasks like on-orbit manipulation, in-situ resource utilization, and on-orbit servicing, where robots must interact closely with other spacecraft or structures. Multi-arm robots can manipulate objects while maintaining their own stability, offering a balance between flexibility and stability.

Among the various robotic configurations, crawling robots with multiple arms have attracted significant attention for their potential in in-space assembly (ISA) tasks \citep{Belvin2016ISA, roa2022pulsar}. These robots can attach themselves to a target spacecraft using their end-effectors, providing stability and enabling precise manipulation. By relocating their base through the motion of their manipulators, crawling robots effectively expand their workspace without the need for large manipulators or excessive fuel consumption \citep{MISHRA2022235}. This paper focuses on multi-arm crawling robot designed for such applications, allowing its end-effectors to dock with a target spacecraft. 



Trajectory optimization (TO) methods have emerged as a powerful tool in the space robotics community for guidance and control. These methods formulate motion planning problems as optimization tasks, considering the robot's dynamics, kinematics, and environmental constraints \citep{apgar2018fast}. Recent trends involve using convex programming to identify locally feasible kinematic paths for spacecraft and free-floating robotic systems \citep{basmadji2020space,l2022trajectory}. Similar approaches have been applied to multi-legged robots, formulating the optimization task as a set of constraints and decision variables to manage complex motions \citep{aceituno2017simultaneous,zhou2023cascade, Jelavic2023LSTP}. Additional constraints specific to the guidance of on-orbit multi-arm robots have been explored to enhance performance \citep{pomarestrajectory}, and to further optimize the interaction forces with the environment during the motion phase \citep{RedondoICMAE24}. A work relevant to our case is that in \citep{Rodriguez2024Aeroconf}, where the authors present a system for multi-armed robots performing on-orbit satellite assembly. It focuses on a higher level task and motion planning (TAMP) setup using a two-layer approach: a high-level logic layer for task sequencing and a low-level layer for trajectory generation via Stochastic Trajectory Optimization (STOMP) \citep{Kalakrishan2011STOMP}. In contrast, our formulation is not tied to a specific task such as on-orbit satellite assembly, but instead remains agnostic to the high-level task being performed. Additionally, our approach integrates low-level control of the arms via reinforcement learning (RL), and accounts for dynamic coupling effects introduced by motion, rather than relying solely on kinematic abstractions.

Despite their effectiveness, traditional TO methods have significant limitations, particularly their reliance on accurate system models and precise environmental representations. Such models are difficult to obtain due to the uncertainties inherent in space missions. Simplifications made during the modeling stage, such as assuming perfect state estimation, known contact states, or flawless execution of planned trajectories, can lead to suboptimal performance or failures in real-world scenarios. Accurately executing these trajectories requires robust control strategies. Model-based control approaches, which rely on predefined mathematical models of the system, are prone to failure due to simplifications and inaccuracies in the models \citep{Kolvenbach2024LeggedSystemsSpace}. They may not adapt well to disturbances or uncertainties not accounted for during the modeling stage.


Reinforcement Learning (RL) has recently emerged as a powerful alternative for developing adaptive and robust control policies without the need for explicit system models. RL algorithms learn control policies through interactions with the environment, allowing them to adapt to disturbances and uncertainties due to the wide variety of scenarios faced during training \citep{sutton2018reinforcement}. They have shown impressive results in complex control problems, including legged locomotion \citep{lee2020,miki2022} and navigation in cluttered environments \citep{rudin2022advancedskills,miki2024learning}. They have also shown potential to learn control policies that can adapt to different physical configurations \citep{belmonte2022metarl}, which is a promising design direction in space robotics.

Despite this success in terrestrial applications, the use of RL-based control in space-related applications is scarce. In this area, RL has primarily been applied to spacecraft guidance, navigation, and control (GNC) tasks. A comprehensive review of RL-based spacecraft GNC systems is presented in \citep{TIPALDI20221}, covering applications such as landing control on celestial bodies, orbital control, and maneuver planning for orbit transfers. Another area of application of RL is planetary robotics, where learning-based controllers have been employed alongside other techniques in order to navigate and explore other celestial bodies in extreme environments, martian-analog or moon-analog terrains \citep{uckert2020investigating, kolvenbach2021traversing, tranzatto2022Cerberus, Arm2023LeggedExplore}. These studies demonstrate the robustness and adaptive capabilities of RL agents in handling system and environment uncertainties.

However, the application of RL techniques to on-orbit servicing and multi-arm robotic control is minimal. \citep{WU2020105657} used an RL-based policy was used to perform position control of two robot manipulators fixed to a satellite in a simplified scenario. In addition, our previous work explores the control of a multi-arm robot in simple trajectories, illustrating the potential of RL to manage the subtleties of on-orbit operations with high-dimensional action spaces \citep{Belmonte2024SPAICE}. 

\begin{figure*}[ht]
    \centering
    \includegraphics[width=\textwidth]{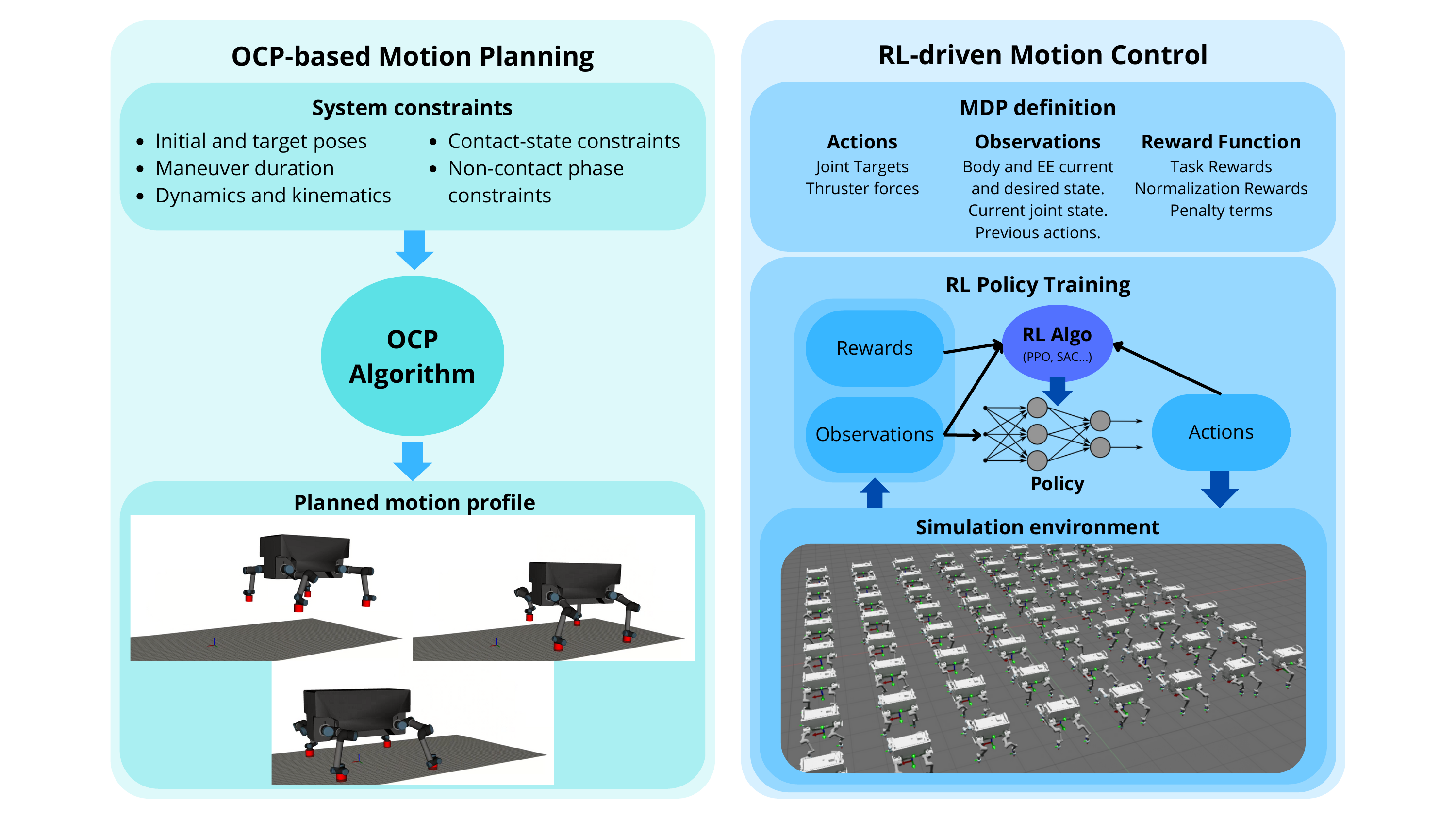}
    \caption{\textbf{Overview of the hybrid approach proposed in this paper.} TO-based motion planning provides a reference motion for the robot with constraints specific to an on-orbit application, while the RL-driven motion control robustly and adaptively tracks this reference motion.}
    \label{fig:overview}
\end{figure*}

To leverage the strengths of both optimization-based planning and learning-based control, this paper proposes a hybrid approach. An overview of our method is depicted in \cref{fig:overview}. The trajectory optimization provides a feasible and efficient path considering the robot's dynamics and environmental constraints, while the RL-driven control method adapts to uncertainties and disturbances during trajectory execution in a model-free fashion. This combination enhances the robot's performance in non-deterministic workspaces and improves its ability to handle uncertainties, while not being computationally expensive at deployment time unlike other machine learning techniques.

In summary, the main contributions of this paper are as follows:
\begin{enumerate}
\item \textbf{Contact-aware trajectory optimization with full orbital dynamics}. We formulate an optimal control problem that jointly optimizes end-effector trajectories, thruster force profiles, and contact timing using polynomial spline parametrization. The model captures complete system dynamics, including base-arm coupling, orbital perturbations, and multi-phase contact interactions, thus enabling smooth, fuel-efficient behavior suitable for microgravity conditions.

\item \textbf{Adaptive timing and task-relevant cost design}. The proposed optimization adjusts phase durations and penalizes tracking errors, contact forces, and robot velocities. This enhances mission safety, reduces fuel consumption, and enables autonomous scheduling of docking and undocking actions.

\item \textbf{High-dimensional policy learning for multi-arm and thruster control}. We develop a reinforcement learning policy over a 27-DoF action space that robustly manages the challenges of orbital manipulation, including coupled dynamics and model mismatch, achieving closed-loop control in one of the most complex settings in the robotics literature.

\item \textbf{Robust RL framework for orbital deployment}. The learning process incorporates orbital-aware reward shaping, curriculum strategies, domain randomization, and observation noise injection to ensure generalization and sim-to-real transfer under realistic spaceflight uncertainties.
\end{enumerate}

The remainder of the paper is organized as follows: \cref{sec:systemarch} introduces the architecture of robotic systems considered for the framework presented in this paper, while \cref{sec:to} describes the main formulation of the trajectory optimization approach and details the system constraints introduced during the optimization. \cref{sec:rl} presents the RL-based motion control strategy, including the design and training of the RL agent. \cref{sec:results} then showcases the results obtained, evaluating the validity and robustness of the proposed approach through simulations and comparisons with existing methods. Finally, \cref{sec:conclusions} summarizes the main findings and discusses the implications of our study for future research and applications in space robotics.

%% file: sections/system_architecture_and_dynamics.tex
\section{System Architecture and Dynamics}
\label{sec:systemarch}

This section details the kinematic and dynamic modeling of a generic multi-arm robot suitable for our framework. The robot consists of $\zeta$ arms, each with $n$ degrees of freedom (DoF). Each arm is equipped with a docking mechanism at its end-effector. The joint coordinates of each arm are represented as $\boldsymbol{q}_i \in \Re^n$ for $i = 1...\zeta$. The robot's coordinate frame, denoted as \textit{B}, is located at the center of its body. The coordinate frame associated with the target spacecraft is referred to as the target coordinate frame \textit{O}. Both frames are orbiting Earth, which serves as the origin of the Earth-centered inertial coordinate frame, denoted \textit{I}. It is assumed that a three-dimensional map of the workspace or the surface of the target spacecraft is constructed, $m(x, y)$, from the point cloud data captured by a camera mounted on the robot's body.

Additionally, thrusters integrated into the robot’s base enable precise maneuvers for reaching and interacting with the target spacecraft surface. These thrusters also facilitate repositioning of the robot's body and compensate for coupled motions induced by the arms and other perturbations.

The frame \textit{B} is rigidly attached to the robot's body, with its origin coinciding with the center of mass $G_B$. The robot's main thruster, denoted $th_{z+}$, is aligned along the body axis  $\boldsymbol{z}_B$ and provides thrust in the positive $z$-direction. The other five small thrusters, namely, $th_{z-}, th_{x+}, th_{x-}, th_{y+}$, and $th_{y-}$, enable maneuvering in other directions. The robot's kinematics are defined by the position and attitude of the robot base, $\boldsymbol{d}_b$ and $\boldsymbol{\phi}_b$ (both relative to the inertial frame), as well as by the joint configuration of each arm, $\boldsymbol{q}_i$.

The complete kinematic state of the multi-arm robot is given by the vector $\left[\boldsymbol{d}_b^T, \boldsymbol{\phi}_b^T, \boldsymbol{q}_i^T \right]^T$.

The control inputs to the robot's body are collected into a vector:
$$
\boldsymbol{u}_{th} = \begin{bmatrix}
    th_{+x} & th_{-x} & th_{+y} & th_{-y} & th_{+z} & th_{-z}
\end{bmatrix}^T
$$
where each element is non-negative (a thruster can only push the spacecraft). The rotation matrix $\prescript{b}{}{\boldsymbol{R}}_I$ represents the robot's attitude relative to the inertial reference frame and can be expressed in terms of quaternions $Q_b = \begin{bmatrix} q_b & \boldsymbol{q}_b \end{bmatrix}^T$ as follows:
\begin{equation}\label{rot_quaternion} \prescript{b}{}{\boldsymbol{R}}_I = \left[(q_b^2 - \boldsymbol{q}_b^T\boldsymbol{q}_b)\boldsymbol{E} + 2\boldsymbol{q}_b\boldsymbol{q}_b^T - 2q_b \boldsymbol{\tilde{q}}_b \right] \end{equation}

Here, $\boldsymbol{E}$ denotes the identity matrix, and $\boldsymbol{\tilde{q}}_b$ is the skew-symmetric matrix formed from the vector part of the quaternion. The kinematic equations describing the robot's body attitude are expressed using the quaternion form:

\begin{equation}\label{eq:kinematic_eq_attitude_quat} \Dot{Q}_b = \frac{1}{2} \boldsymbol{\Omega}(\boldsymbol{\omega}_b)Q_b = \frac{1}{2}\begin{bmatrix} -\boldsymbol{\omega}_b^T & 0 \\ \boldsymbol{\tilde{\omega}}_b & \boldsymbol{\omega}_b \end{bmatrix} \begin{bmatrix} \boldsymbol{q}_b \\ q_b \end{bmatrix}
\end{equation}

where $\boldsymbol{\omega}_b$ represents the angular velocity of the robot's body. Rigid body dynamics is used to model the robot's body motion due to the end-effector contact forces and thrusters forces:
\begin{equation}\label{eq:rigidbodydyn1} m\Ddot{\boldsymbol{d}}_b= \sum_1^\zeta \left[\boldsymbol{f}_i(t)\right]+\boldsymbol{f}_b
\end{equation}
\begin{equation}\label{eq:rigidbodydyn2}
\begin{split}
\boldsymbol{I}_i\Dot{\boldsymbol{\omega}}_b(t)+\boldsymbol{\omega}_b(t) \times \boldsymbol{I}_i\boldsymbol{\omega}_b(t)= \\ \sum_1^\zeta \boldsymbol{f}_i(t)\times (\boldsymbol{d}_b-\boldsymbol{p}_{i}(t)) + \boldsymbol{I}_{i}^{-1} \boldsymbol{T}_e
\end{split}
\end{equation}
where the mass of the entire robot is given by $m$, $\boldsymbol{f}_i(t)$ is the external wrench action of each arm, $\boldsymbol{I}_i$ is a constant rotational moment of inertia calculated from the nominal robot configuration, the manipulators end-effector positions are represented by $\boldsymbol{p}_i$, the linear forces $\boldsymbol{f}_b$ are generated from the robot base thrusters, and $\boldsymbol{T}_e$ represents external disturbing torques acting on the satellite, such as gravity gradient torque.

\begin{figure}
    \centering
    \includegraphics[width=\columnwidth]{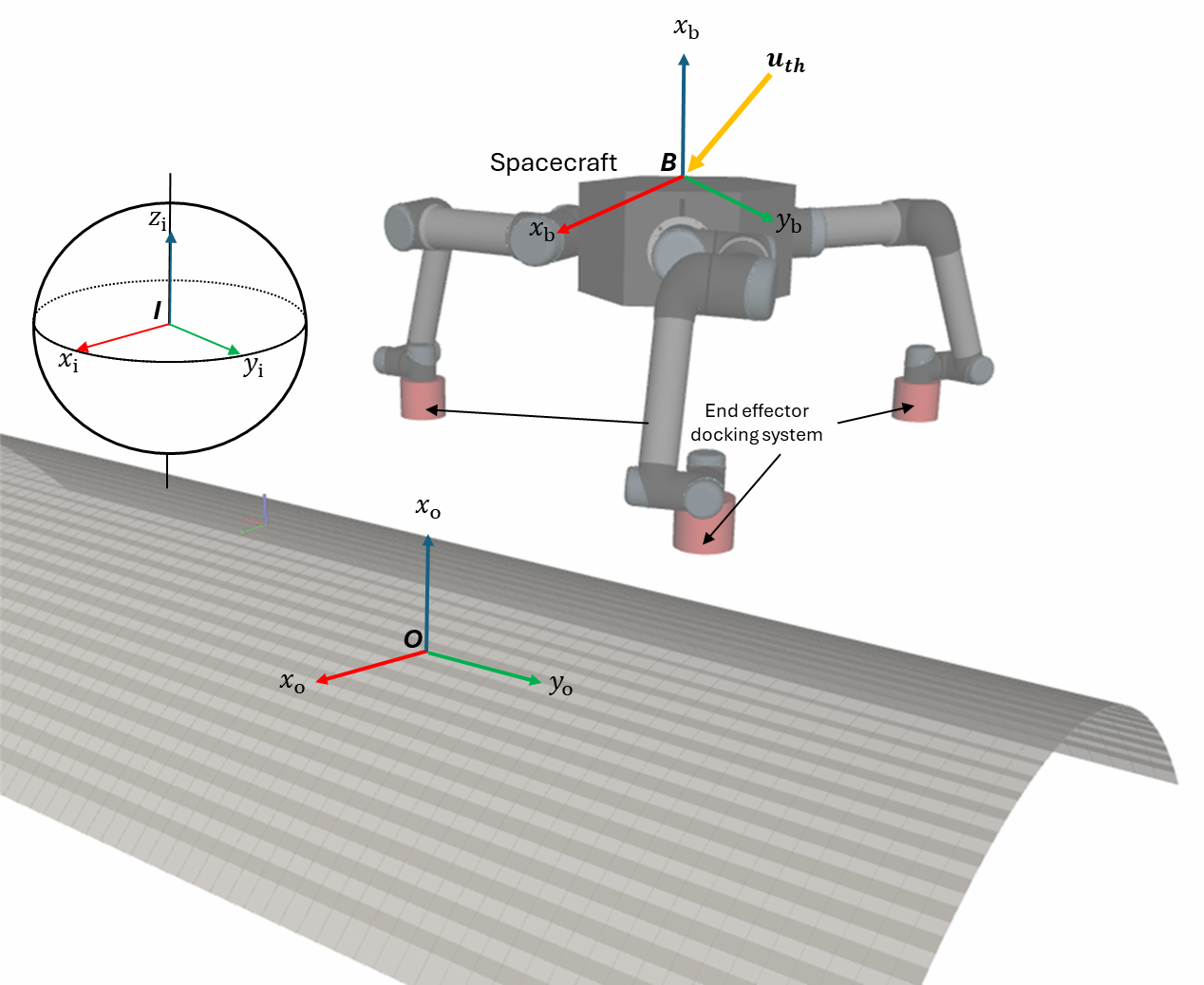}
    \caption{\bf{Visualization of the described system architecture and on-orbit operational environment.}}
    \label{fig:system_architecture}
\end{figure}

To illustrate the described scenario, \cref{fig:system_architecture} provides a visualization of an example application with a 3-arm robotic system. The different parts of the robot and coordinate systems are also included for clarity. Note that, while the image shows a 3-arm robot, our method is generic and can be applied to an arbitrary multi-arm platform.

%% file: sections/trajectory_optimization.tex
\section{Trajectory Optimization}\label{sec:to}

The path planning for the multi-arm robot is generated using a trajectory optimization (TO) approach based on formulating an Optimal Control Problem (OCP). The proposed OCP formulation extends the work presented in \citep{RedondoICMAE24} and is described in detail in this section.

Considering $\boldsymbol{d}_b$ as the robot linear center of mass (CoM) and $\boldsymbol{\phi}_b$ as its orientation, the information required by the TO algorithm is the initial robot body location $\left[\boldsymbol{d}_b(t=0) = \boldsymbol{d}_{b0}, \boldsymbol{\phi}_b(t=0) = \boldsymbol{\phi}_{b0} \right]$, the desired final robot body location $\left[\boldsymbol{d}_b(t=T) = \boldsymbol{d}_{bd}, \boldsymbol{\phi}_b(t=T) = \boldsymbol{\phi}_{bd} \right]$, the number of contact phases for each arm, $N_i$, and the total duration of the maneuver, $T$. The OCP automatically generates the robot body trajectory, $\left[\boldsymbol{d}_b(t), \boldsymbol{\phi}_b(t) \right]$, the control actions applied to the robot body, $\boldsymbol{u}_{th}$, the manipulators end-effector trajectories, $\boldsymbol{p}_i(t)$, the interaction forces of each arm end-effector with respect to the contact surface, $\boldsymbol{f}_i(t)$, and the required gait for each arm during the trajectory.

The base spacecraft thrusters can be represented by linear forces with respect to the body center of mass $\boldsymbol{u}\in \Re^3$. We construct a continuous spline by combining multiple third-order polynomials and optimize the coefficients of these polynomials to achieve the desired properties. The thruster force profiles are  parametrized using $N_b$ segments. We employ three polynomials of equal duration $T_{b,j}/3, j=1...N_b$ for each segment $N_{b,j}$. The duration of each segment, and, consequently, the duration of each thruster's polynomial is adjusted since $T_{b,j}/3$ is included as one of the decision variables in the OCP. This parametrization is capable of capturing commonly varying thruster force profiles while maintaining the problem's minimal complexity. Therefore, alternating sequences of cubic polynomials, $u(t)=a_0+a_1t+a_2t^2+a_3t^3$ are considered. Considering $u_0, \Dot{u_0}, u_1, \Dot{u_1}$ the value and first derivative at the beginning and end of a polynomial with duration $T_{b,jn}$, the coefficients are obtained by considering: 
\begin{equation}\label{hermit}
\begin{aligned}
a_0 &= u_0, \\
a_1 &= \Dot{u}_0, \\
a_2 &= -T_{b,jn}^{-2} \big[3(u_0 - u_1) + T_{b,jn}(2\Dot{u}_0 - \Dot{u}_1)\big], \\
a_3 &= -T_{b,jn}^{-3} \big[2(u_0 - u_1) + T_{b,jn}(\Dot{u}_0 - \Dot{u}_1)\big].
\end{aligned}
\end{equation}
Furthermore, the conditions used as the end of the  previous polynomial can also be used as the starting node of the  next, which ensures continuous force changes over the trajectory. We predefine the maximum  number of segments $N_b$. It is worth noting that this is not a significant limitation, as segment durations can be reduced to nearly zero if they are unnecessary. However, the algorithm retains the flexibility to adjust the phase durations. Given that these durations are variable, it is essential to ensure that the total duration of each force spline aligns with the specified overall time, $T$, and therefore, $\sum_{j=1}^{N_b} T_{b,j} = T$.

A similar parametrization is used for robot end-effector trajectories and interaction forces. However, each arm's trajectory comprises two types of phases: the contact phases $(t \in C_i)$ and the non-contact trajectory phases $(t \notin C_i)$. The durations of these phases for arm $i$ are denoted as $T_{i,j}$, where $j = 1 \ldots 2N_i$, and they satisfy $\sum_{j=1}^{2N_i} T_{i,j} = T$. The OCP automatically determines the durations of these phases, which define the gait of the arms. Consequently, the manipulators' end-effector trajectories and interaction forces are partitioned into these phases as $\boldsymbol{p}_i (t, T_{i,1}, \ldots)$ and $\boldsymbol{f}_i (t, T_{i,1}, \ldots)$, respectively. In the same way as for the thruster forces, these trajectories are encoded using different polynomials of fixed durations that are joined to create a continuous spline whose coefficients are optimized. For instance, for a given arm's trajectory $\boldsymbol{p}_i(t \notin C_i)$, multiple third-order polynomials are considered for each non-contact phase, and a constant value is used during the contact phases. Conversely, for each arm's force profile $\boldsymbol{f}_i(t \in C_i)$, multiple polynomials represent each contact phase, and zero force is established during the non-contact phases. The duration of each phase, and therefore the duration of each arm's polynomial, is adjusted based on the optimized phase durations. Analogously, the OCP simultaneously optimizes the six-dimensional trajectory of the robot base, which includes the linear path $\boldsymbol{d}_b(t)$, with the associated orientation expressed using Euler angles $\boldsymbol{\phi}_b(t)$. 

Various constraints are defined to ensure realistic motions by guaranteeing the robot's kinematic and dynamic properties. The kinematic constraint ensures that the range of motion for each arm $i$, is consistent with the robot's kinematics. This constraint is defined as a prism with edge length $2\xi_i$, centered at the nominal position for each arm $i$:
\begin{equation}\label{eq:kinem_eom}
    |\boldsymbol{R}_b [\boldsymbol{p}_i(t) - \boldsymbol{d}_b^T(t)] - \boldsymbol{p}_{ni}| < \xi_i
\end{equation}
where $\boldsymbol{p}_{ni}$ is the nominal position for each arm's end-effector, and $\boldsymbol{R}_b$ represents the attitude of the robot base with respect to the inertial frame (rotation matrix). This kinematic constraint ensures the allowable range of movement for each arm while simultaneously avoiding self-collisions by restricting the arms to non-overlapping regions. Once the range of motion is guaranteed, dynamic constraints are also considered to obtain realistic motions. 

Additional constraints are enforced during contact phases $t \in C_i$ to ensure stable docking positions throughout each phase. Moreover, the $z$ coordinate of the contact points should correspond to the 3D map $m(x, y)$, which is assumed to be obtained by a camera mounted on the robot; that is, $p_{ic}^{z}(t \in C_i) = m(p_i^x, p_i^y)$. This 3D map provides information about the depth or $z$ component of the surface at coordinates $(x, y)$ with respect to frame $B$. Additionally, to prevent the arm's end-effector from slipping during the contact phase, the following constraint is included: $\boldsymbol{\Dot{p}}_i(t \in C_i) = \boldsymbol{0}$, ensuring that the docking position is maintained.




Regarding the non-contact phases, further constraints are necessary to ensure safe motion. First, no contact forces should be generated when $t \notin C_i$, as the arm will be in the swing phase. Therefore, the following constraint is included during these phases: $\boldsymbol{f}_i(t \notin C_i) = \boldsymbol{0}$. Second, a collision avoidance constraint is included to maintain a minimum distance $\delta$ between the robot body and the target surface. This constraint is expressed as:
\begin{equation}\label{eq:height_delta}
    p_{ic}^{z}(t) - m(p_i^x, p_i^y) > \delta
\end{equation}
With these constraints, the main considerations added to the OCP formulation have been described. An overview of all the system constraints is provided in \cref{tab:constraints}. 

\begin{table}[t]
\small \sf \centering
\caption{\bf System Contraints for the OCP}
\label{tab:constraints}
\begin{tabular}{c|c}
\toprule
\bfseries \textbf{Constraint} & \bfseries \textbf{Value} \\
\midrule
Starting robot pose & $\left[\boldsymbol{d}_{b0}, \boldsymbol{\phi}_{b0} \right]$ \\
Target robot pose & $\left[\boldsymbol{d}_{bd}, \boldsymbol{\phi}_{bd} \right]$ \\
Maneuver duration & $\sum_{j=1}^{N_i} \Delta T_{i,j} = T$ \\
Dynamics and Kinematics & \cref{eq:kinem_eom,eq:rigidbodydyn1,eq:rigidbodydyn2} \\
\hline
\textbf{Contact phase} $(t \in C_i)$ & \\
\hline
End-effector is fixed & $\boldsymbol{\Dot{p}}_i(t \in C_i) = \boldsymbol{0}$ \\
Docking on target & $p_{ic}^{z}(t \in C_i) = m(p_i^x, p_i^y)$ \\
Avoid collisions & \cref{eq:height_delta} \\
\hline
\textbf{Non-contact phase} $(t \notin C_i)$ & \\
\hline
Null interaction force & $\boldsymbol{f}_i(t \notin C_i) = \boldsymbol{0}$ \\
\hline
Collision avoidance & $p_{ic}^{z}(t) - m(p_i^x, p_i^y) > \delta$  \\
\bottomrule
\end{tabular}
\end{table}

Finally, the function to be optimized by the OCP algorithm is given by:
\begin{equation}\label{eq:ocp_cost_function}
\begin{split}
    \Phi = \int_0^T \sum_1^\zeta \left[\sigma_{i1}\left(f_i^x(t) \right)^2 + \sigma_{i2}\left(f_i^y(t) \right)^2 + \sigma_{i3}\left(f_i^z(t) \right)^2 \right] + \\
    \sigma_4\left(\Dot{d}_b^x(t) \right)^2 + \sigma_5\left(\Dot{d}_b^y(t) \right)^2 + \sigma_6\left(\Dot{d}_b^z(t) \right)^2 dt
\end{split}
\end{equation}
where $f_i^x(t)$, $f_i^y(t)$, and $f_i^z(t)$ are the contact forces at the manipulators' end-effectors in the $x$, $y$, and $z$ directions; $\dot{d}_b^x(t)$, $\dot{d}_b^y(t)$, and $\dot{d}_b^z(t)$ are the velocities of the robot base in each direction; and $\sigma_i$ are weighting constants. These terms are included in the cost function to minimize the required contact forces and to obtain smooth trajectories for the robot body.

The optimization framework explicitly addresses the physical interaction with the environment in two key ways:

Firstly, the kinematic constraints strictly enforce consistency between the base motion and the fixed end-effector positions. This ensures that the robot's generated motion does not violate geometric constraints, preventing excessive stress or "tearing" forces at the contact points even during single-arm anchoring.

Secondly, in multi-contact phases where the system forms a closed kinematic chain, redundant forces (internal forces) could theoretically arise. Our cost function \cref{eq:ocp_cost_function} minimizes the squared norm of contact forces $\sum ||\boldsymbol{f}_i||^2$. This naturally penalizes antagonistic force components that do not contribute to the centroidal motion, effectively resolving the redundancy and minimizing internal stress on the structure.

The pseudocode for the optimization process is summarized in Algorithm \ref{alg:ocp_alg}. As indicated, the continuous OCP is transcribed into a finite-dimensional Nonlinear Programming (NLP) problem via phase-based parameterization and temporal discretization. The decision variables include the polynomial coefficients for the base motion, end-effector trajectories, and contact forces, as well as the phase durations. The resulting NLP is solved using the Interior Point Optimizer (IPOPT), a primal-dual interior-point method designed for large-scale non-linear optimization. To ensure computational efficiency and robust convergence, the gradients (Jacobians) and Hessians of the objective function and constraints are computed via analytical derivatives. This avoids numerical approximation errors and significantly speeds up the solver iterations compared to finite-difference methods. The average computation time for the full multi-arm trajectories presented in Section 5 is lower than 10 seconds on a standard desktop workstation. While this currently supports offline planning for the RL training pipeline, the formulation is compatible with real-time requirements if adapted to a Receding Horizon Control (RHC) scheme with lower-frequency updates and warm-starting.

In the proposed tasks, the robot operates in moderately structured environments, such as locomotion over varied surfaces or performing approaches to these surfaces using thrusters. It is assumed that the workspace is not highly cluttered and that the robot does not need to navigate through narrow or severely constrained spaces. Under these conditions, the reduced-dimensionality, phase-based parameterization enables efficient and reliable planning. For more complex environments with high obstacle density or occlusions, the trajectory optimization method can be integrated with global planners or multi-start optimization techniques, which represent promising directions for future work.

\begin{algorithm}[t]
    \caption{Trajectory Optimization (OCP) for Multi-Arm Robot}
    \label{alg:ocp_alg}
    \textbf{Input:} Initial state $d_b(0), \phi_b(0)$; Target body pose $d_b(T), \phi_b(T)$; Maneuver duration $T$; Target spacecraft map $m(x,y)$; Number of contact segments $N_i$; Number of thruster segments $N_b$; cost weights $\sigma$.
    
    \textbf{Output:} Optimized trajectories, thrusters and contact forces $[d_b(t), \phi_b(t), u_{th}(t), p_1(t), ..., p_{\zeta}(t), f_1(t), ..., f_{\zeta}(t)]$.
   
    \begin{algorithmic}[1]
        \STATE \textbf{Parametrization:}
        \STATE \quad Construct spline-based trajectories for base, manipulators, thrusters and contact forces; 
        \STATE \quad Define segment durations $T_{b,j}$ and $T_{i,j}$, and spline coefficients as decision variables.
        
        \STATE \textbf{Minimize Cost Function:}
        \STATE \quad $\Phi = \int_{0}^{T} \sum_{i=1}^{\zeta} [\sigma_{i1}(f_{i}^{x})^2 + \sigma_{i2}(f_{i}^{y})^2 + \sigma_{i3}(f_{i}^{z})^2] \, dt + \int_{0}^{T} [\sigma_{4}(\dot{d}_{b}^{x})^2 + \sigma_{5}(\dot{d}_{b}^{y})^2 + \sigma_{6}(\dot{d}_{b}^{z})^2] \, dt$ 
        
        \STATE \textbf{Subject to Constraints:}
        
        \STATE \quad Total duration consistency: $\sum_{j=1}^{2N_i} T_{i,j} = T$ and $\sum_{j=1}^{N_b} T_{b,j} = T$
        \STATE \quad Dynamics: $\dot{d}_b, \dot{\omega}_b$ defined by Eqs. 3, 4.
        \STATE \quad Kinematic Limits: $|R_{b}[p_{i}(t)-d_{b}^{T}(t)]-p_{ni}|<\xi_{i}$
        
        \STATE \quad For Contact Phases ($t \in C_i$):
        \STATE \quad \quad Fixed Position: $\dot{p}_{i}(t) = 0$ 
        \STATE \quad \quad Docking on Map: $p_{ic}^{z}(t) = m(p_{i}^{x}, p_{i}^{y})$ 
        \STATE \quad \quad Force Continuity: $f_i(t)$ are continuous splines
        \STATE \quad For Non-Contact Phases ($t \notin C_i$):
        \STATE \quad \quad Null Force: $f_{i}(t) = 0$
        \STATE \quad \quad Collision Avoidance: $p_{ic}^{z}(t)-m(p_{i}^{x},p_{i}^{y})>\delta$
        
        \STATE \textbf{Solve} the Non-Linear Programming (NLP) problem using IPOPT.
        \STATE \textbf{Return} Optimized trajectories, thrusters and contact forces.
    \end{algorithmic}
\end{algorithm}

%% file: sections/rl_framework.tex
\section{Reinforcement Learning-driven Control}\label{sec:rl}

Reinforcement learning (RL) has emerged as one of the most powerful approaches towards robust and adaptive robot control \citep{tang2024deepreinforcementlearningrobotics}. Despite specific nuances, every RL application requires modeling the task as a Markov Decision Process (MDP) \citep{lauri2023MDP}. 

An MDP is defined by the tuple $(\mathcal{S}, \mathcal{A}, \mathcal{P}, \mathcal{R})$, where $\mathcal{S}$ is the state space describing the task at hand, $\mathcal{A}$ is the action space that the agent can apply to the environment, ${\mathcal{P}}(s_{t+1} | s_t, a_t)$ is the probability density function describing the transition probabilities between states, and ${\mathcal{R}}(s_t, a_t, s_{t+1}): \mathcal{S} \times \mathcal{A} \times \mathcal{S} \rightarrow \Re$ is the reward function describing the task at hand. However, in some cases, the state is not fully observable, and thus we only have access to a subset of the full state description $s_t$. This is called a Partially Observable Markov Decision Process (POMDP), which is the most common case in robotic tasks since we cannot typically retrieve the whole state of the robot, but only the information available from its sensors. To manage such cases, we introduce an observation space $\mathcal{O}$, from which we will obtain an observation $o$ which is usually a subset of its corresponding state $s$, and is sampled from a distribution $o \sim \mathcal{G}(o_t | s_t, a_t, s_{t+1})$.

At each timestep, the agent observes the environment and obtains an observation vector $o_t \in \mathcal{O}$ that partially describes the current state $s_t \in \mathcal{S}$. It then takes an action $a_t \in \mathcal{A}$ based on its policy $\pi_{\theta}(a_t | o_t)$, transitions to a new state $s_{t+1}$ based on $\mathcal{P}$, and receives a scalar reward signal $r_t(s_t, a_t, s_{t+1})$. The final goal of the RL framework is to adjust the parameters $\theta$ of the policy $\pi_{\theta}$ to obtain an optimal policy $\pi_{\theta}^*$ which maximizes the cumulative discounted rewards:

\begin{equation}\label{eq:max_cum_rew}
    \pi_{\theta}^* = \underset{\theta} 
 {\mathrm{argmax}}{\mathbb{E}}\left[\sum_{t=i}^{\infty} \gamma^t r_t\right]
\end{equation}

where $\gamma \in (0,1) $ is the discount factor that modulates the importance of prioritizing short-term or long-term rewards. The adjustment of the parameters $\theta$ and the subsequent optimization of the policy is achieved through an iterative process that consists of sequential interactions with the environment, in a "trial and error" loop, which resembles the way most animals learn. 

\begin{figure*}[ht]
    \centering
    \includegraphics[width=\textwidth]{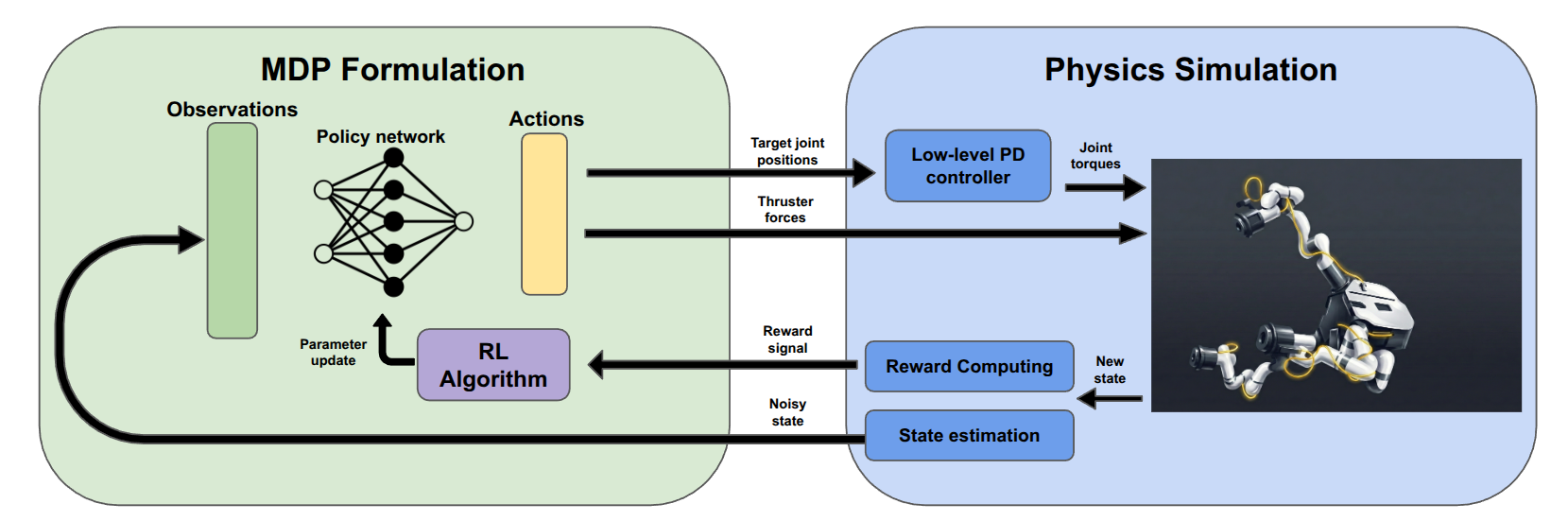}
    \caption{\textbf{Overview of the Reinforement Learning-Driven control methodology}. The policy network receives as input an observed robot state, and outputs desired joint positions and thruster forces for the robot. These are fed to the physics simulation, yielding a new state. The transition is used to update the network parameters via an RL algorithm of choice, and the loop is repeated until convergence.}
    \label{fig:RLOverview}
\end{figure*}

While this is the general framework for sequential decision-making problems we need to characterize the MDP for our desired task, namely tracking a pre-planned trajectory while adapting the motion to possible disturbances or inaccuracies in the generated path. The description of the MDP and the training environment for our task is an extension of our previous work \citep{Belmonte2024SPAICE}. An overview of the proposed MDP formulation and general RL workflow is illustrated in \cref{fig:RLOverview}, and described in detail as follows.

The action space $\mathcal{A}$ consists of two main action groups: The first involves the target joint positions for each leg, $\boldsymbol{q}_i^\ast$. These target joint positions are then converted into joint torques by a PD controller running at higher frequency, as is common in most successful applications of RL for robotic control \citep{lee2020,miki2024learning,arm2024pedipulate,jenelten2024dtc}. Since the policy runs at lower frequency than standard torque controllers, relying on higher-frequency PD controllers ensures that the actuators receive a control signal at a suitable frequency, reducing oscillating behavior. The PD gains are tuned together with the reward terms, since they have a direct effect on policy performance by actively conditioning the system's behavior.

The second action group consists of the applied thruster force at the robot's CoM, $\boldsymbol{u}_{th}$. For simplicity, we decided to group the thruster forces in the main cartesian axis, since two thrusters acting in opposite directions are unlikely to activate at the same time.

The observation space $\mathcal{O}$ represents the information that our policy will use to infer its current state and decide on its next actions based on it. In our setup, it comprises the current base pose and velocities, namely $[\boldsymbol{d}_{b}, \boldsymbol{\phi}_{b}, \boldsymbol{v}_b, \boldsymbol{\omega}_b]$, the desired base pose at each timestep, $[\boldsymbol{d}_{b}^{\ast}, \boldsymbol{\phi}_{b}^{\ast}]$, the joint positions, $\boldsymbol{q}$, and velocities, $\boldsymbol{\Dot{q}}$, the actions outputted by the policy at the previous timestep, $\boldsymbol{a}_{t-1} = [\boldsymbol{q}_{t-1}^{\ast}, \boldsymbol{u}_{th, t-1}]$, and the current and desired end-effector poses at each timestep, noted by $\boldsymbol{p}_i$ and $\boldsymbol{p}_{i}^{\ast}$, respectively. 

\cref{tab:observations} provides an overview of the different terms introduced above for reference. Overall, we include the robot's current and desired states, policy actions, and actuator magnitudes (joint positions and velocities) as observations. These elements, which drive the system's motion, provide the policy with sufficient information to evaluate its distance from the target and effectively learn a mapping from specific observations to actions that yield higher rewards.

\begin{table}[t]
\small \sf \centering
\caption{\bf Policy Observations}
\label{tab:observations}
\begin{tabular}{c|c}
\toprule
\bfseries \textbf{Observation name} & \bfseries \textbf{Expression} \\
\midrule
Base pose and velocities & $\left[\boldsymbol{d}_{b}, \boldsymbol{\phi}_{b}, \boldsymbol{v}_b, \boldsymbol{\omega}_b \right]$ \\
Desired base pose & $\left[\boldsymbol{d}_{b}^{\ast}, \boldsymbol{\phi}_{b}^{\ast} \right]$ \\
Joint positions and velocities & $\left[\boldsymbol{q}, \boldsymbol{\Dot{q}}\right]$\\
Previous actions & $\boldsymbol{a}_{t-1} = [\boldsymbol{q}_{t-1}^{\ast}, \boldsymbol{u}_{th, t-1}]$ \\
Current end-effector pose & $\boldsymbol{p}_i$\\
Desired end-effector pose & $\boldsymbol{p}_{i}^{\ast}$\\
\bottomrule
\end{tabular}
\end{table}

As for the reward function $\mathcal{R}$, we separate the reward terms in three groups: Task rewards $R_{task}$, which describe the main desired behavior to be achieved by our policy, normalization rewards $R_{norm}$ that modulate the policy actions to perform softer and realistic motions, and penalty rewards $R_{pen}$ aiming to avoid undesired states such as collisions with the environment or self-collisions. The total reward function designed for this work will then be ${\mathcal{R}} = R_{task} + R_{norm} + R_{pen}$. As is common, each of the employed terms is weighted using a factor $w$, which serves both to normalize the values employed for computing rewards (i.e., velocities and accelerations are usually of different orders of magnitude), and grant different importance to each of the terms, giving priority to some parts above others.

The task reward, $R_{task}$, is defined in order to track both the desired base pose and end-effector targets, so that $R_{task} = R_{body} + R_{ee}$. To this end, we employ a logarithmic function of the error between the current and desired states, getting higher gradient values when the error is lower, resulting in a more fine-grained tracking performance. The body pose tracking reward is computed as follows:
\begin{equation}
    R_{body} = w_{body} \cdot \left(-ln(||\boldsymbol{d}_{b}^{\ast} - \boldsymbol{d}_{b}|| + \epsilon) - ln(||\boldsymbol{\Psi}|| +\epsilon) \right)
\end{equation}
where $\epsilon$ is a regularization value to ensure the logarithmic function is well defined, and $\boldsymbol{\Psi}$ is the axis angle representation of the rotation error between the desired orientation and the current orientation, computed as the multiplication of the quaternion representing the desired orientation, and the conjugate of the quaternion for the current orientation. The arms' end-effector tracking reward follows an identical form:
\begin{equation}
    R_{ee} = w_{ee} \cdot \sum_{i=0}^{\zeta} \left(-ln(||\boldsymbol{p}_{i}^{\ast} - \boldsymbol{p}_{i}|| + \epsilon) - ln(||\boldsymbol{\psi}_i|| +\epsilon) \right)
\end{equation}
where $\psi_i$ is the analog of $\Psi$ for each of the $\zeta$ arms. Since both the body pose and end-effector poses need to be tracked accurately, we need to be cautious while setting the weighting factors for each term in order to not prioritize one of the terms much more than the other. 

For the normalization reward $R_{norm}$, the focus is on reducing jerky moves and excessive energy consumption, resulting in smoother motions for both the arms and the body, while also being more energetically efficient. To achieve this, we discourage high values for joint power, joint accelerations, body acceleration, and difference between consecutive joint targets (to avoid oscillations). We also penalize heavy usage of the body thrusters to prevent high fuel consumption, as well as relying entirely on them to perform the desired maneuver. Taking this into account, the resulting reward term is formulated as follows:
\begin{multline} 
    R_{norm} = w_{pow} \cdot \sum_i (\dot{\boldsymbol{q}}_{i}^{T} \boldsymbol{\tau}_{i})^2 + w_{acc} \cdot \sum_i \Ddot{\boldsymbol{q}}_i^2 \\+ w_{bacc} \cdot (||\Dot{\boldsymbol{v}}_b|| + ||\Dot{\boldsymbol{\omega}}_b||) + w_{act} \cdot \sum_i (\boldsymbol{q}_{i,t}^{\ast} - \boldsymbol{q}_{i, t-1}^{\ast})^2 \\ + w_{th} \cdot ||\boldsymbol{u}_{th}||
\end{multline}

where all symbols have already been defined across the paper.

The final penalty term, $R_{pen}$, heavily penalizes actions leading to collisions with the environment or with the robot itself. In an on-orbit servicing scenario, even the smallest collision can lead to losing grip of the target spacecraft, or dealing irreparable damage to the robot. Thus, we decided to take a conservative approach in this regard, and heavily penalize these situations during training, such that only the arms' end-effectors can establish contact with any surrounding surface. With this, the penalty reward term is defined as: 
\begin{equation}
    R_{pen} = w_{c} \cdot n_c + w_{bc} \cdot b_c
\end{equation}
where $n_c$ is the number of contacts of the arms with any surface that is not the robot body, and $b_c \in \{0,1\}$, with $b_c = 1$ denotes termination due to a body collision.

The last part of the MDP definition is the modelling of the agent's policy itself. We parametrize the policy $\pi_{\theta}(a_t | o_t)$ as a Multi-Layer Perceptron (MLP) similar to most works on RL-based robot control, since MLPs are high-capacity universal function approximators ideal to represent the dynamics of a robotic system. We employ an stochastic Gaussian MLP with three hidden layers of 512 neurons each and an ELU activation function \citep{elu}, with similar hyperparameters to those in previous works \citep{rudin2022advancedskills}.

%% file: sections/results.tex
\section{Results}\label{sec:results}

This section presents simulation experiments to showcase the behavior of the two main contributions of this work: The extended TO formulation to generate feasible motion patterns for multi-arm robots using both the arms and thrusters to reach a desired configuration, and the RL-driven model-free control policy that is able to robustly follow the planned trajectory.

\subsection{Experimental Setup}\label{sec:setup}

\subsubsection{System and tasks description.} \label{sec:setup_description}
To illustrate the performance of our approach, we analyze two different case studies. In the first case, the robot is already in contact with the target spacecraft and needs to traverse across the spacecraft surface. We present two different tasks for our system in this scenario.

In the first task, the robot must achieve a displacement of 1.2m in the $x$ direction, with the orientation and motion in other directions remaining unchanged, within a target time of $T=20s$. For the second task, it needs to reach a target position displaced both in the $x$ and $y$ directions. Specifically, our system has to move 1.5m in $x$ direction, while shifting -0.5m in $y$ direction within a target time of $T=25s$, maintaining the same body orientation. With this second experiment, we aim to show how a combined forward and sideways motion is carried out by our proposed framework.

Furthermore, we perform an empirical evaluation of how the number of time-segmented polynomials affects the motion profile generated by the TO algorithm in order to validate the selection made for our case studies.

For the second case, the robot starts with its end effectors at a height of 0.5 m with respect to the surface of the spacecraft, and must employ the thrusters to reach the surface and then move towards the desired target pose, which is 1.2m forward in the $x$ direction. This maneuver must be completed within a target time of $T=20s$.

For all the experiments, we will employ the robotic system depicted in \cref{fig:tako}. The choice of four arms provides the necessary redundancy and stability for complex ISA tasks. For this work, we consider that every point in the target spacecraft is a possible docking point. We do so because the focus of our work relies on the improvements in planning and control strategies, rather than on precise docking operations. That said, specific docking locations can also be considered by using the proposed approach. Both the TO algorithm and RL policy run directly on the robotic system onboard computer, meaning that no delays or latencies affect the planning and control stages. The desired body and end effector positions would be provided by a human operator or a higher-level task planning algorithm such as that described in \cite{Rodriguez2024Aeroconf}.

The mass of the system is $m=250kg$, considering the robot body, the four arms, and the fuel employed by the thrusters. The main kinematics and dynamic parameters of the robot arms are detailed in the Appendix. The robot consists of $\zeta=4$ arms, with $n=6$ DoF each. This results in an action space of dimension $4 \times 6 + 3 = 27$, which to the best of our knowledge is one of the highest dimensions explored in learning-based robotic control. To provide some context, \cite{Peng2018DeepMimic} shows Deep RL-based control of physics-based characters for specific skills, including a 34 DoF humanoid robot. \cite{OpenAiFiveFinger} reports control of a 24 DoF dexterous robotic hand. \cite{Radosavic2024Humanoid} performs control of a humanoid robot of 26 DoF by using a teacher-student RL framework.

\subsubsection{Implementation details.} \label{sec:rl_training}

It is worth mentioning that the OCP-based TO method explored here utilizes a simplified dynamics model, which considers that the entirety of the robot's mass is located at its CoM (see \cref{eq:rigidbodydyn1,eq:rigidbodydyn2}). While this suffices to solve the planning problem, it introduces certain nuances that the RL-driven controller will need to manage in order to bridge the gap between the simplified model and the full-dynamics system.

\begin{figure}[t]
    \centering 
    \includegraphics[width=\columnwidth]{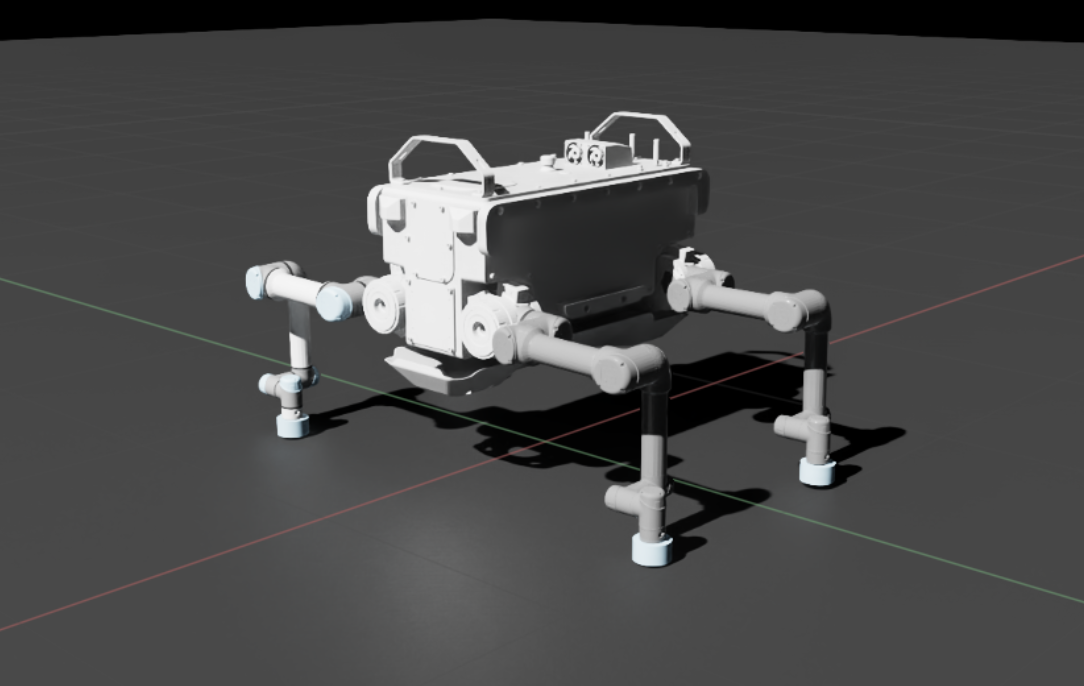}
    \caption{\textbf{Multi-arm robot used in our experiments.}} 
    \label{fig:tako}
\end{figure}

Regarding the training details of the RL-based control policy, we use NVIDIA's \textit{Isaac Sim} \citep{makoviychuk2021isaacgymhighperformance} as the high-performance simulator to recreate our on-orbit servicing scenario, and the \textit{Orbit} robot learning framework (currently Isaac Lab) to define the training environment \citep{mittal2023orbit}. While the TO algorithm operates with the simplified centroidal dynamic model mentioned earlier, here we employ a full-dynamics, realistic USD model for our multi-arm robot (see \cref{fig:tako}). Thus, the RL policy will need to adapt a motion planned for a body with centroidal dynamics to a system with mass distributed across the robot and the inertias and coupled motions introduced by the arms' motions.

We implement the observation and action spaces described in \cref{sec:rl}, as well as the reward function, to define the training environment for our agent. The specific weights for each of the reward terms described in \cref{sec:rl} are included in the Appendix.

For policy training, instead of generating trajectories during training and forwarding the planned motion as in \citep{jenelten2024dtc}, we chose to learn a policy to reach a desired base and end-effector pose which is "close" to the current robot state. This decision was taken to heavily reduce computational costs. Each trajectory optimization run takes about $10s$ on our training machine, so creating thousands of different trajectories at every episode would become computationally prohibitive. In addition, if we only generated a set of trajectories and use them for training, the resulting policy could be highly biased towards those pre-planned trajectories, which is also not desirable. 

We hypothesize that training a policy capable of reaching a nearby desired base and end-effector poses, sequentially feeding the targets generated by the planner at deployment time, would also allow the robot to follow the trajectory, as well as permitting other interesting applications such as positioning a single arm in a desired target to perform a servicing activity.  

To achieve our pose tracking goal, we follow an approach of increasing the difficulty of the task sequentially with a target distance curriculum. We start by fixing the target base pose to the initial pose, and only sample targets for the end-effector positions in a sphere of $0.2m$ around the initial end-effector position, with the target changing twice per episode to provoke learning of quick end-effector target variations. In this way, the policy will first learn to maintain the stability of the base while moving the arms to a desired configuration. After a certain threshold, we start expanding the range of desired base poses, increasing distances as long as the policy is able to reduce the base pose error below a defined threshold of $5cm$. This is, every time that the average base pose error is below $5cm$, the range of sampled target base positions is increased by $10cm$.

Lastly, to improve policy robustness to model mismatch and uncertain state estimation, we perform domain randomization during training, and inject noise into some of the policy observations described in \cref{sec:rl}. Specifically, we randomize the initial robot and joint positions within certain boundaries, and also randomize the system mass by a maximum of $\pm 10\%$ at each episode reset to simulate different initial fuel conditions. For noise injection, we add slight noise sampled from a uniform distribution to the values that would be obtained using noisy sensor measurements (body position and orientation, linear and angular velocities, and joint measurements). The specific values employed for domain randomization and noise injection are included in \cref{tab:domain_randomization} and \cref{tab:noise} in the Appendix. As \cref{sec:systemarch} describes, we assume that thrusters do not generate residual torques, simplifying their modeling as purely force-producing actuators. While it is difficult to perfectly align thrusters to avoid torque generation in real spacecraft systems, the trained RL policy is expected to handle such small residual torques due to its general capability to compensate for reaction forces and torques induced by arm motions. Domain randomization and noise injection techniques enhance the policy’s ability to handle unmodeled disturbances, including small residual torques from thruster misalignments.

Each training episode lasts $15s$ if it is not terminated earlier due to a collision, and at each reset the initial arm joint positions are randomized within a $\pm 25\%$ of their default value. The policy and the PD controller run at $50Hz$ and $200Hz$ respectively, following the work in \citep{jenelten2024dtc,Belmonte2024SPAICE}.

We use PPO \citep{schulman2017proximal} as the training algorithm, with the same hyperparameters as in previous works \citep{Belmonte2024SPAICE}. We train the policy with 8192 parallel environments and a batch size of $24 \times 8192$, which results in 15h for 25000 epochs on an NVIDIA GeForce RTX 3090Ti. More details of the hyperparameters used can be found in the Appendix.

\subsection{Case 1: Motion across the spacecraft surface.}
\subsubsection{Displacement of 1.2 meters in $x$ direction}
\paragraph{Trajectory Optimization:}\label{sec:TO_exp}
We begin with the experiment in which the robot moves to a target pose on the spacecraft, being  initially in contact. In order to showcase the improvements made with the formulation presented in \cref{sec:to}, we compare the trajectory obtained with our current formulation to computing the same motion profile with the TO algorithm introduced in our previous work \citep{RedondoICMAE24}, which lacks the body thrusters that allow robot's motion to be stabilized. 

\begin{figure*}[ht!]
    \centering
    \begin{subfigure}{0.49\textwidth}
         \centering
         \includegraphics[width=\textwidth]{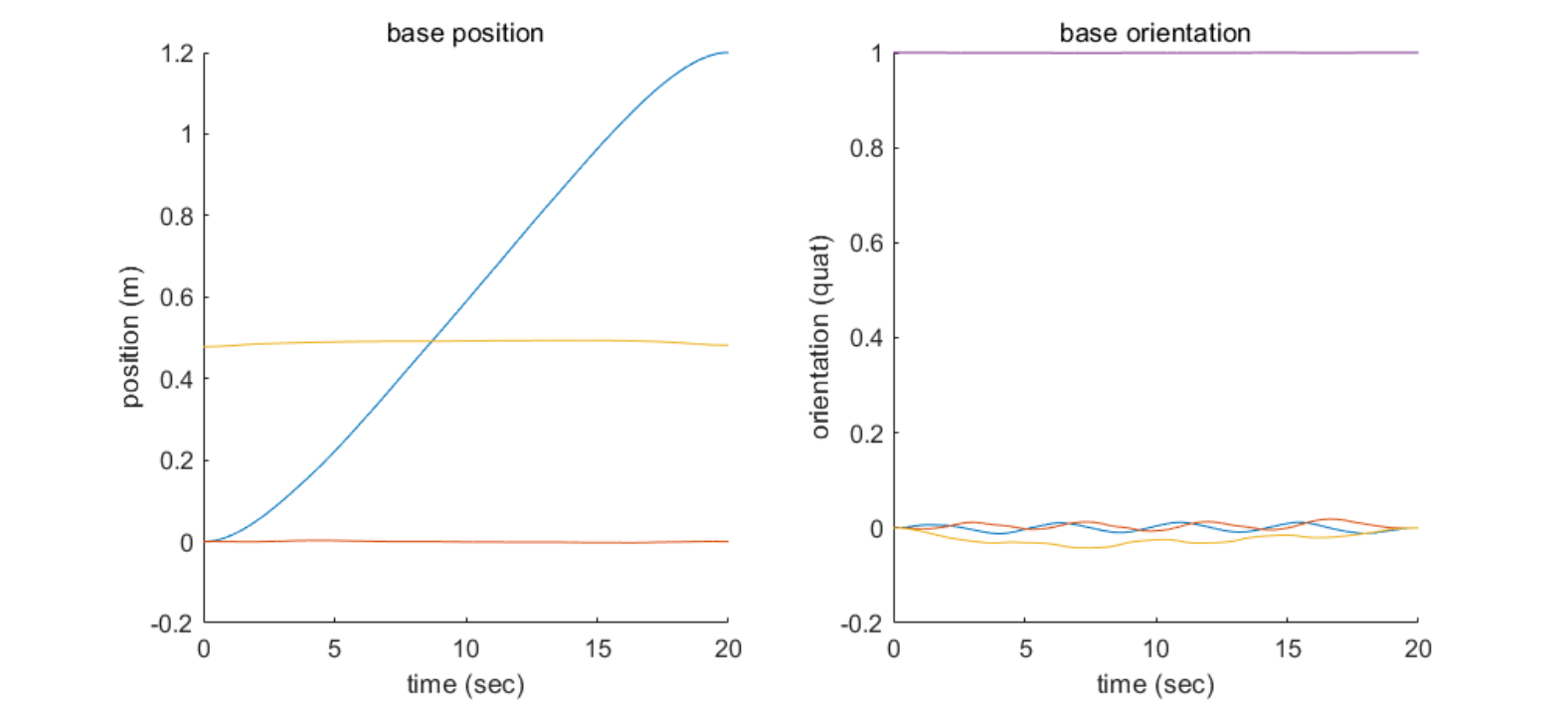}
         \caption{With thrusters}
         \label{fig:body_th}
    \end{subfigure}
    \begin{subfigure}{0.49\textwidth}
         \centering
         \includegraphics[width=\textwidth]{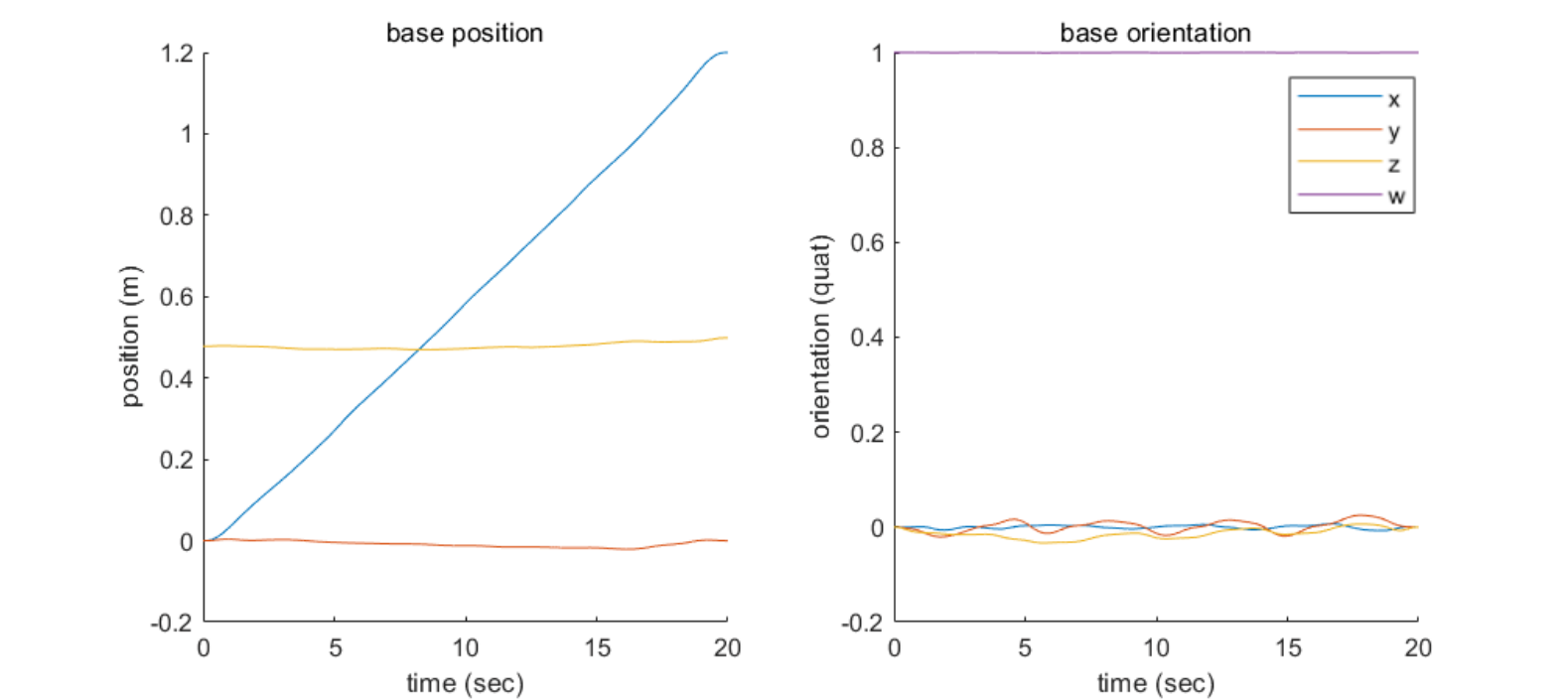}
         \caption{Without thrusters}
         \label{fig:body_noth}
    \end{subfigure}    
\caption{\bf{Position and orientation of the robot base during the trajectory}}
\label{fig:body_traj}
\end{figure*}

\begin{figure*}[ht!]
    \centering 
    \begin{subfigure}{0.4\textwidth}
        \centering
        \includegraphics[width=\textwidth]{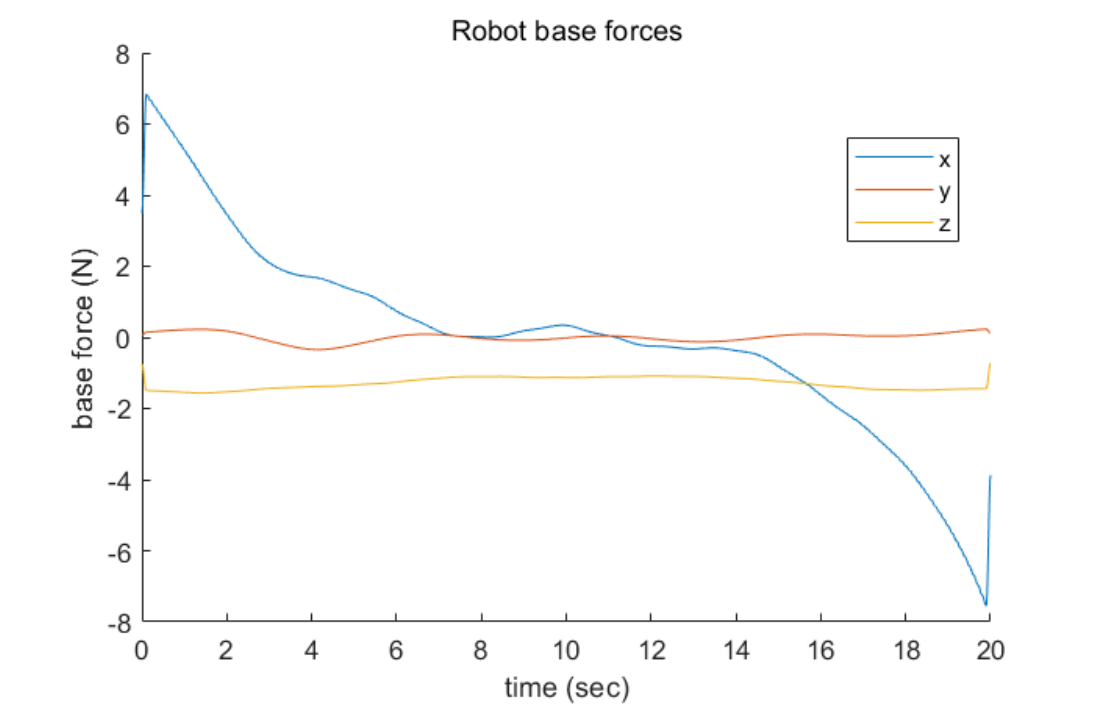}
        \caption{\bf{Robot base thruster forces during the trajectory.}} 
        \label{fig:thruster}
    \end{subfigure}
    \begin{subfigure}{0.59\textwidth}
        \centering
        \includegraphics[width=\textwidth]{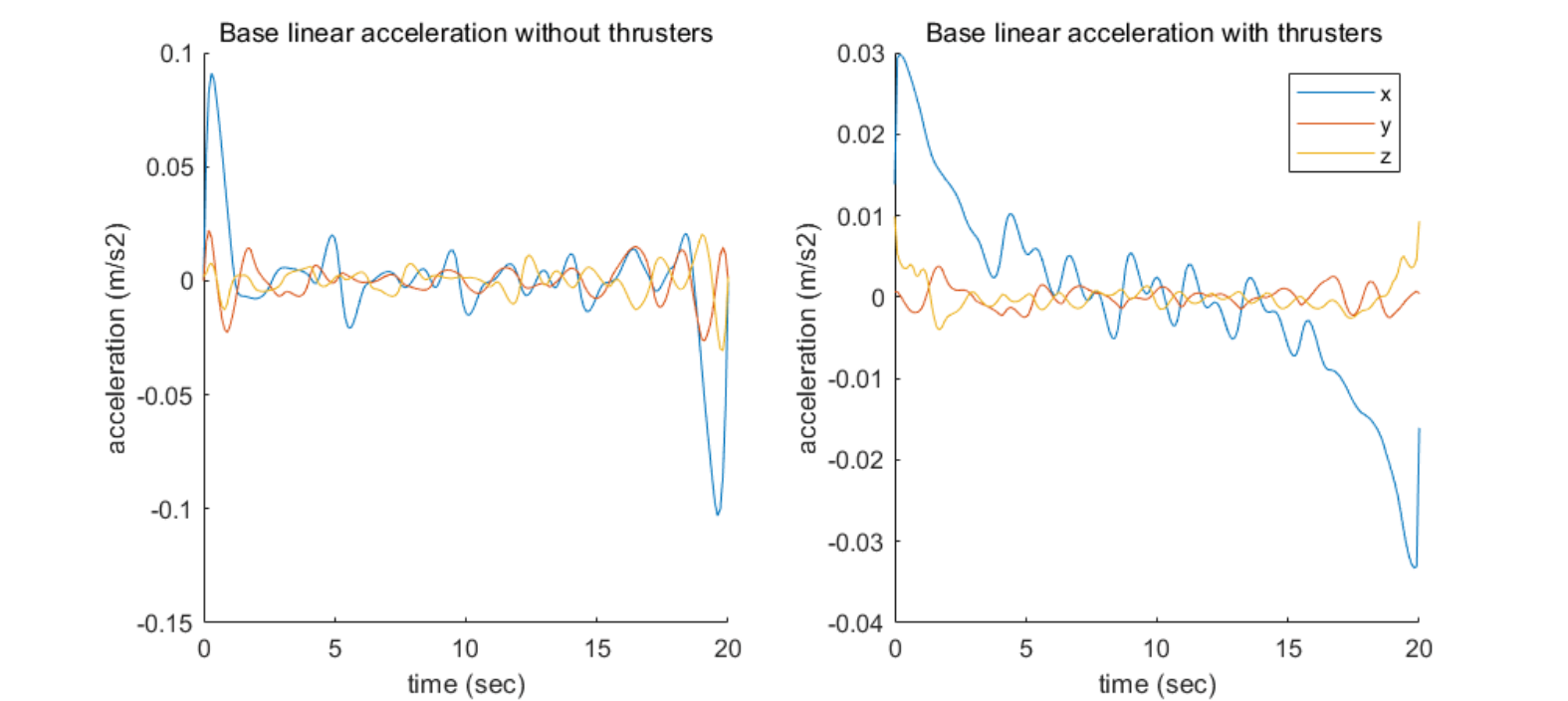}
        \caption{\bf{Robot base accelerations during the trajectory}}
        \label{fig:comacc}
    \end{subfigure}
    \caption{\bf{Robot base thruster force and acceleration with and without thrusters}}
\end{figure*}

With these considerations, the TO method computes the desired body trajectory, the target motion profile for the end-effectors, the desired interaction forces when in contact with the surface, and the thruster forces to be applied to the body. \cref{fig:body_traj} shows the evolution of the position and orientation of the body during the trajectory, both with and without thruster usage. As can be seen, the use of thrusters improves smoothness and stability during motion with respect to the no-thruster case, with practically no oscillations in the base trajectory as shown in \cref{fig:body_th}.

The thruster forces computed by the planner that help to obtain such a smooth motion are depicted in \cref{fig:thruster}. It can be seen that they primarily assist in initiating the motion, and then in decreasing velocity at the end when the target position is getting closer. In addition, the forces in $z$-axis help maintain the initial robot height, and forces in $y$-axis prevent the robot from drifting.

Furthermore, we can also see a considerable improvement in terms of the accelerations that the robot base achieves, as depicted in \cref{fig:comacc}. The use of thrusters helps modulate the acceleration of the body, promoting a smoother motion compared to the accelerations caused by moving entirely on the basis of the accelerations induced by the arms alone. 

\begin{figure*}[ht]
    \centering
    \begin{subfigure}{0.49\textwidth}
         \centering
         \includegraphics[width=\textwidth]{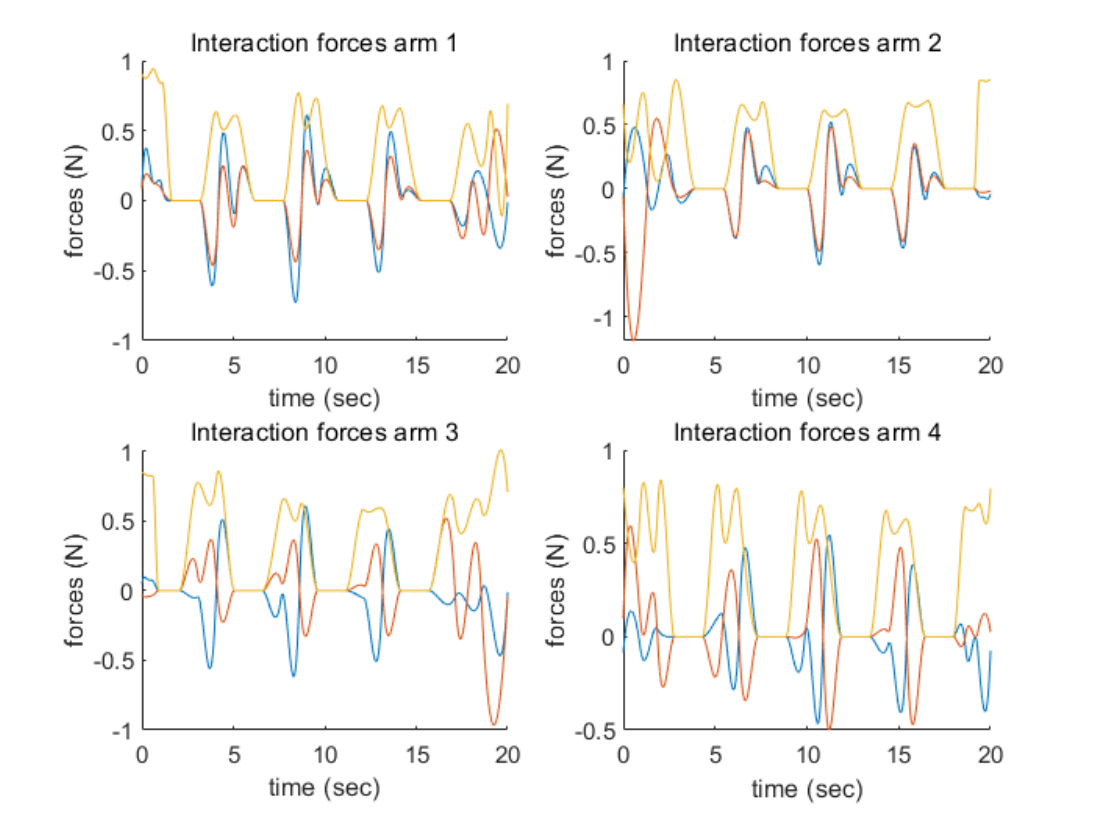}
         \caption{With thrusters}
         \label{fig:forces_th}
    \end{subfigure}
    \begin{subfigure}{0.49\textwidth}
         \centering
         \includegraphics[width=\textwidth]{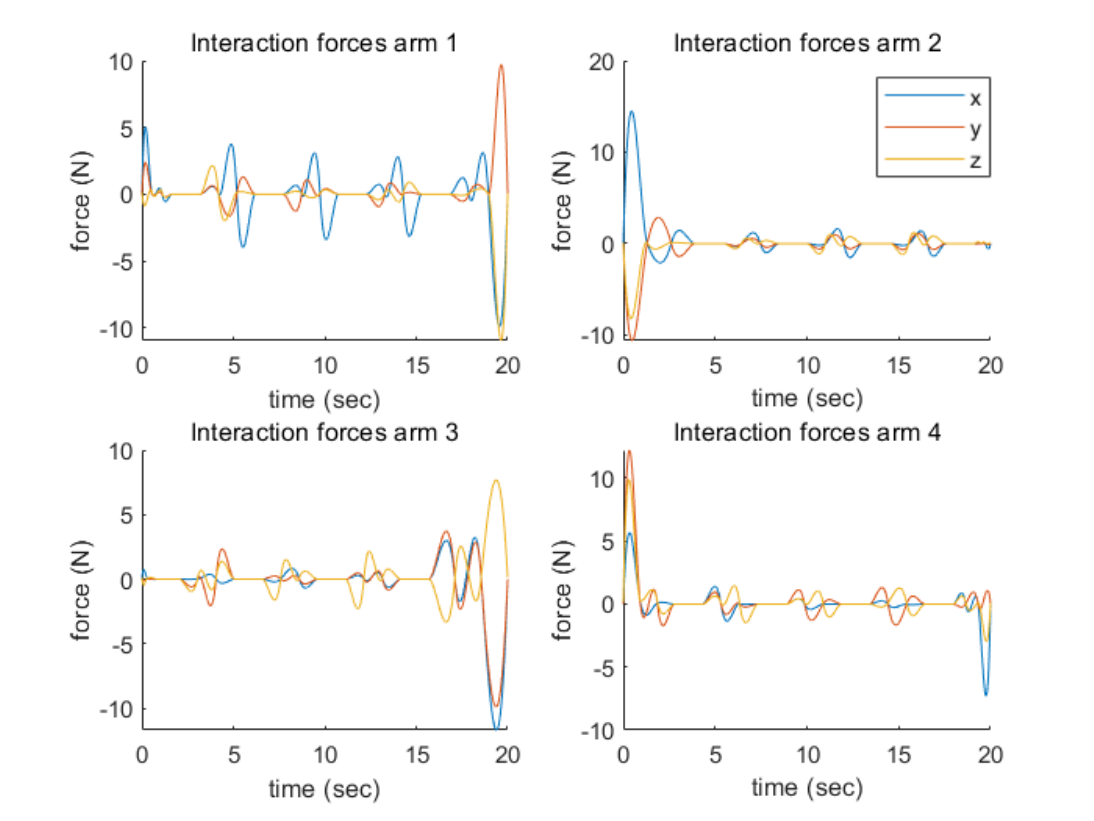}
         \caption{Without thrusters}
         \label{fig:forces_noth}
    \end{subfigure}    
\caption{\bf{End-effector interaction forces}}
\label{fig:forces}
\end{figure*}

However, the greatest impact of thruster usage is shown in \cref{fig:forces}. Here, it can be seen that the slight impulse provided by the thrusters in \cref{fig:thruster} drastically reduces the interaction forces generated at the docking points, especially during the initiation and completion of the maneuver.

\begin{figure*}[ht]
    \centering
    \begin{subfigure}{0.49\textwidth}
         \centering
         \includegraphics[width=\textwidth]{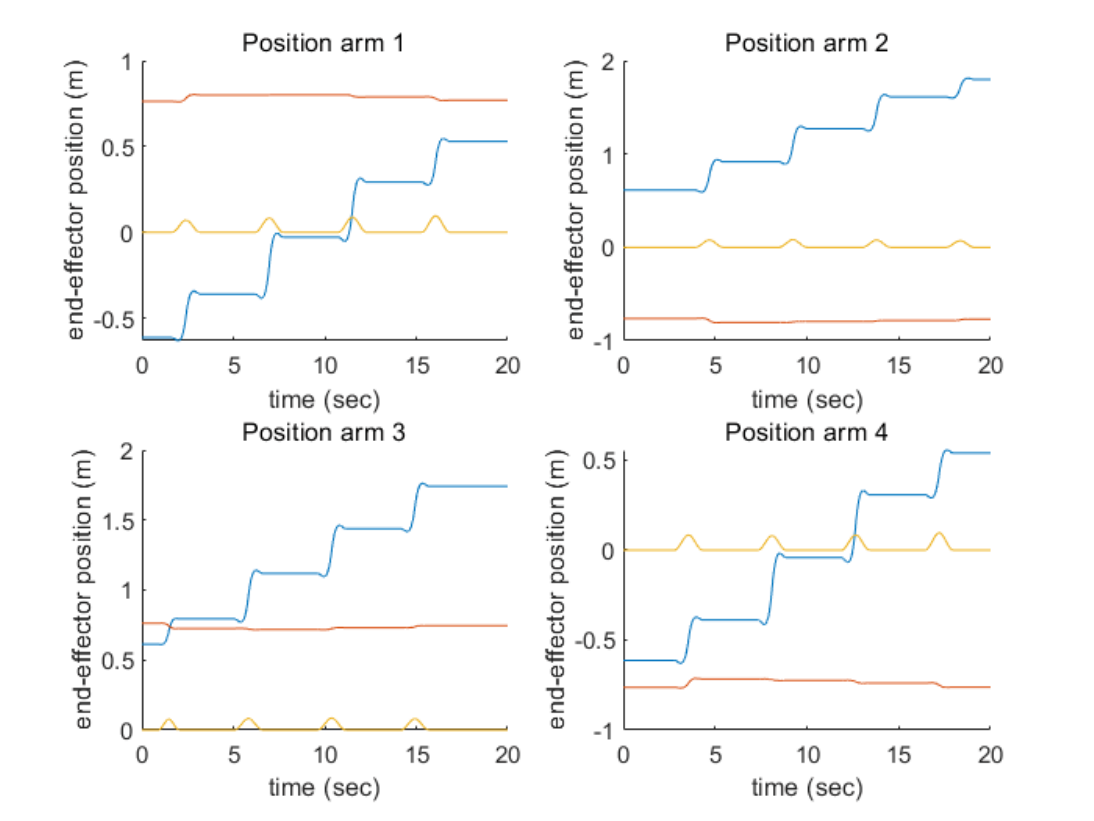}
         \caption{With thrusters}
         \label{fig:ee_th}
    \end{subfigure}
    \begin{subfigure}{0.49\textwidth}
         \centering
         \includegraphics[width=\textwidth]{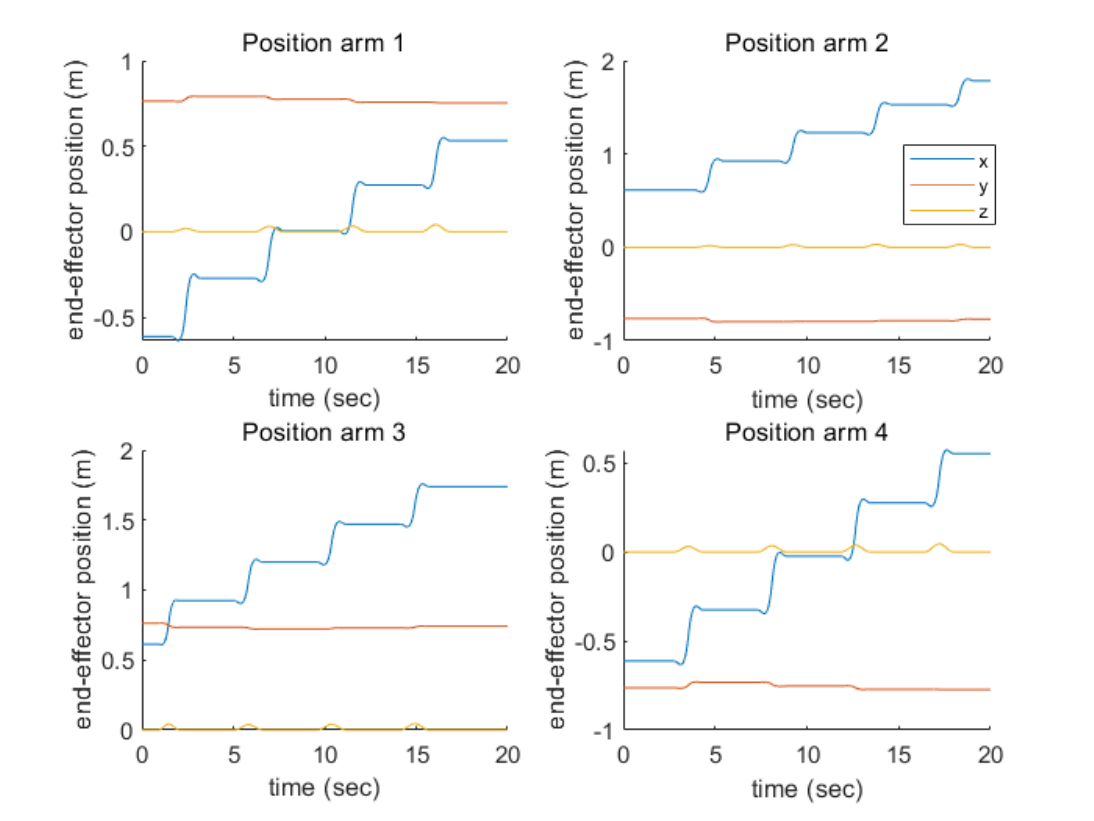}
         \caption{Without thrusters}
         \label{fig:ee_noth}
    \end{subfigure}    
\caption{\bf{End-effector position during the maneuver}}
\label{fig:ee_traj}
\end{figure*}

Finally, the evolution in the end-effector positions is illustrated in \cref{fig:ee_traj}. Here, the difference between contact and non-contact phases discussed in \cref{sec:to} can be clearly seen. The robot moves its arms in the desired direction in such a way that there are always three arms in the contact phase and one in the swing phase, providing more safety and stability throughout the maneuver.

\paragraph{RL-driven Motion Control: }\label{sec:RL_exp}
Having demonstrated the effectiveness of the TO approach, particularly improvements achieved through thruster force integration, we now assess the tracking of this planned trajectory with an RL-driven control policy. 

\begin{figure*}[ht]
    \centering
    \begin{subfigure}{0.32\textwidth}
         \centering
         \includegraphics[width=\textwidth]{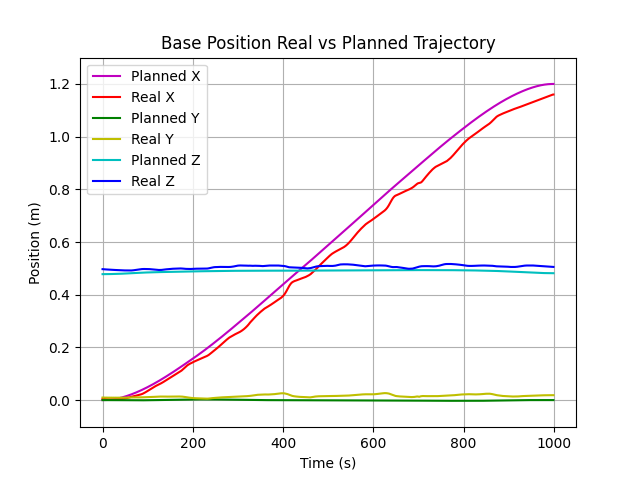}
         \includegraphics[width=\textwidth]{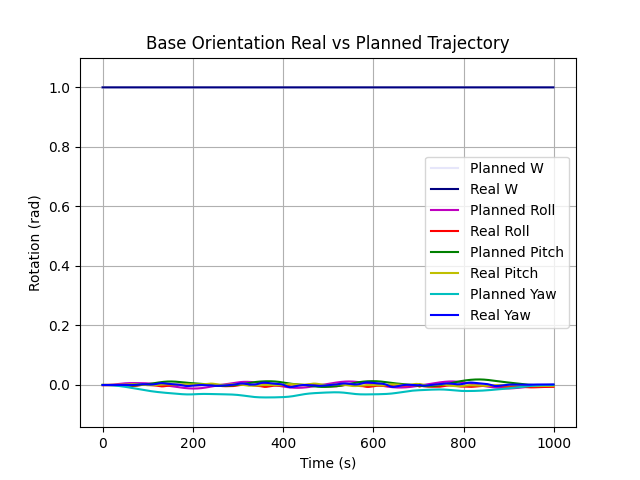}
         \caption{Case 1.1: Body Pose Tracking performance }
         \label{fig:track_body}
    \end{subfigure}    
    \begin{subfigure}{0.67\textwidth}
         \centering
         \includegraphics[width=0.48\textwidth]{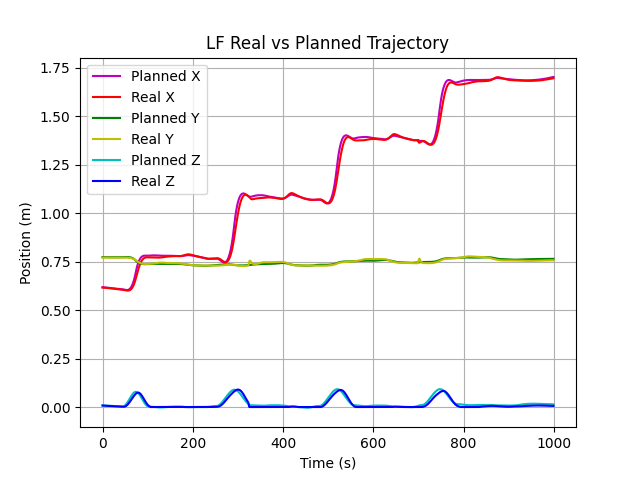}
         \includegraphics[width=0.48\textwidth]{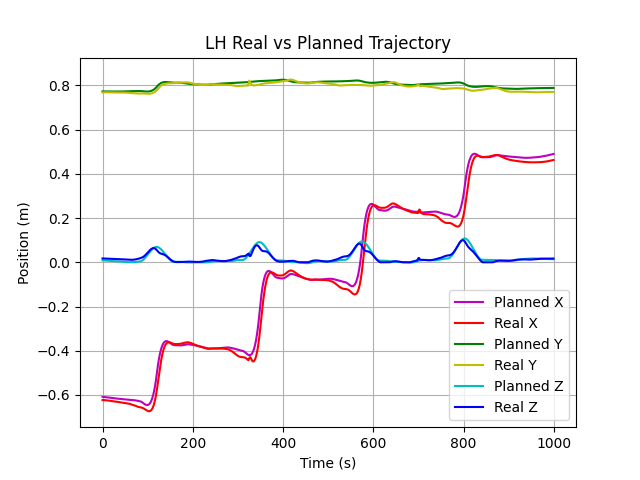}
         \includegraphics[width=0.48\textwidth]{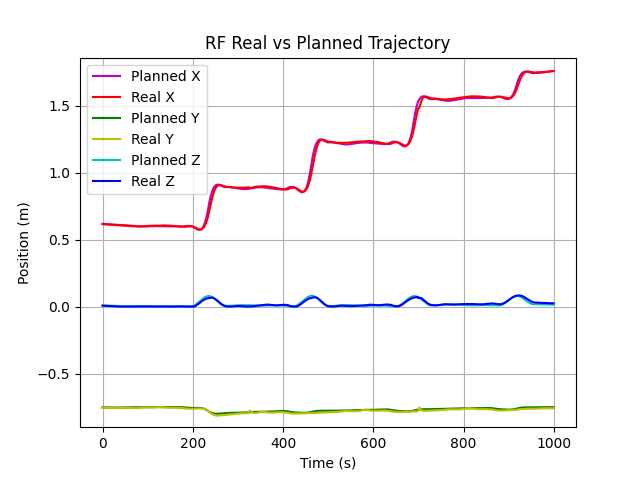}
         \includegraphics[width=0.48\textwidth]{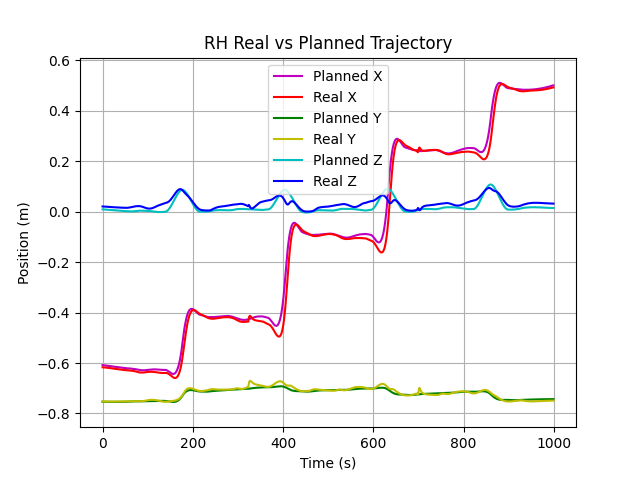}
         \caption{Case 1.1: Arm end-effector tracking performance}
         \label{fig:track_ee}
    \end{subfigure}
\caption{\bf{Case 1.1: Tracking performance using the RL-driven control policy}}
\label{fig:tracking}
\end{figure*}

Once we have trained our control policy, we deploy it for a trajectory tracking task. We employ the trajectory generated in \cref{sec:TO_exp} which also uses thrusters for base stabilization. The TO algorithm provides 100 trajectory points per second, resulting in a total of 2000 target points, with a time difference of $\Delta 0.01s$ each. Since our policy runs at $50Hz$, we subsample the reference trajectory to get targets with a delay of $\Delta 0.02s$ between them, and update the target base pose and end-effector pose at every policy step. \cref{fig:track_body} shows, the body pose trajectory tracking performance of our RL policy. As can be seen, the policy closely follows the reference trajectory in both position and orientation, with a small delay in tracking the reference caused by the mismatch between the moment where the target changes and when the policy actually reaches it.

\cref{fig:track_ee} depicts the performance of the policy in tracking the desired end-effector motion profiles. It can also be seen that the arms follow closely the gait pattern in the trajectory, particularly for the front arms, while the behavior is a little noisier for the right-hind arm. Nonetheless, the overall tracking performance for both the body and the end-effector poses is high, with the RL-driven policy being capable of closely following the reference trajectory generated by the OCP solver, accounting for the full dynamics of the system and not only a simplified model.

\subsubsection{Displacement of 1.5 meters in $x$ direction and -0.5m in $y$ direction.} \label{sec:xy_traj} After evaluating our framework with the forward motion presented in \cref{sec:RL_exp} and validating its performance, we now seek to test our approach for a more complex task of a combined forward and sideways motion profile. This task presents additional challenges such as different reaction forces generated by side motions of the arms, and also introduces the possibility of using the thrusters to assist in the motion for an additional axis.

We generate the trajectory using our TO algorithm as before, resulting in a total of 2500 target points for the $T=25s$ maneuver. As for the previous task, we subsample the reference trajectory to obtain a delay between target points that matches the policy frequency. For the sake of briefness, we omit the plotting of the planned motion and force profiles for this task, and show only the tracking performance of the RL-driven control, since it also includes the tracked motion profiles.

\begin{figure*}[ht]
    \centering
    \begin{subfigure}{0.32\textwidth}
         \centering
         \includegraphics[width=\textwidth]{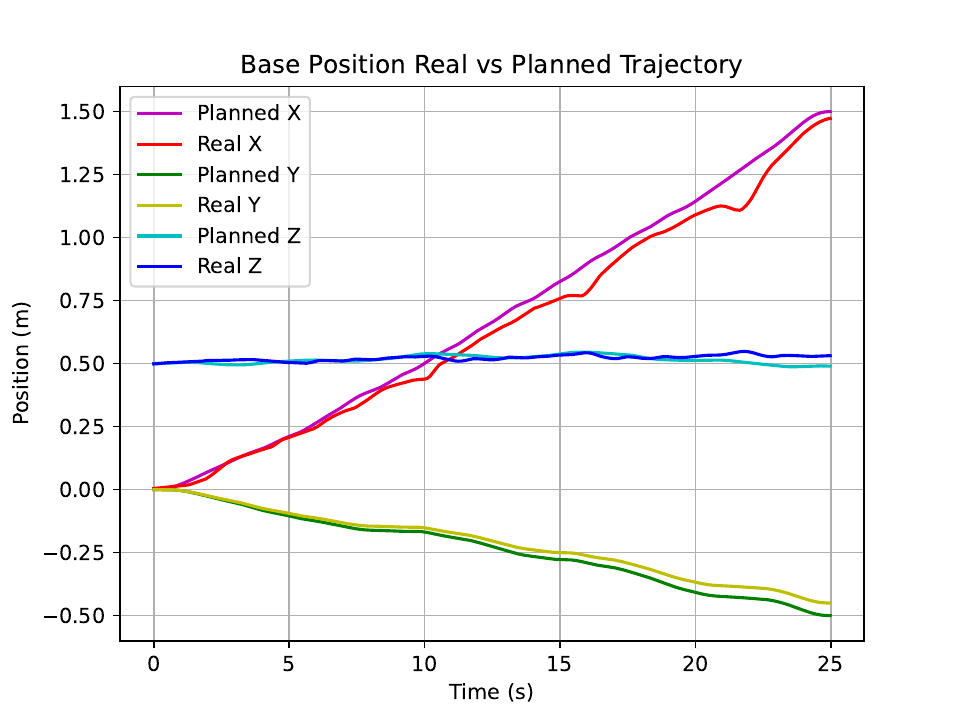}
         \includegraphics[width=\textwidth]{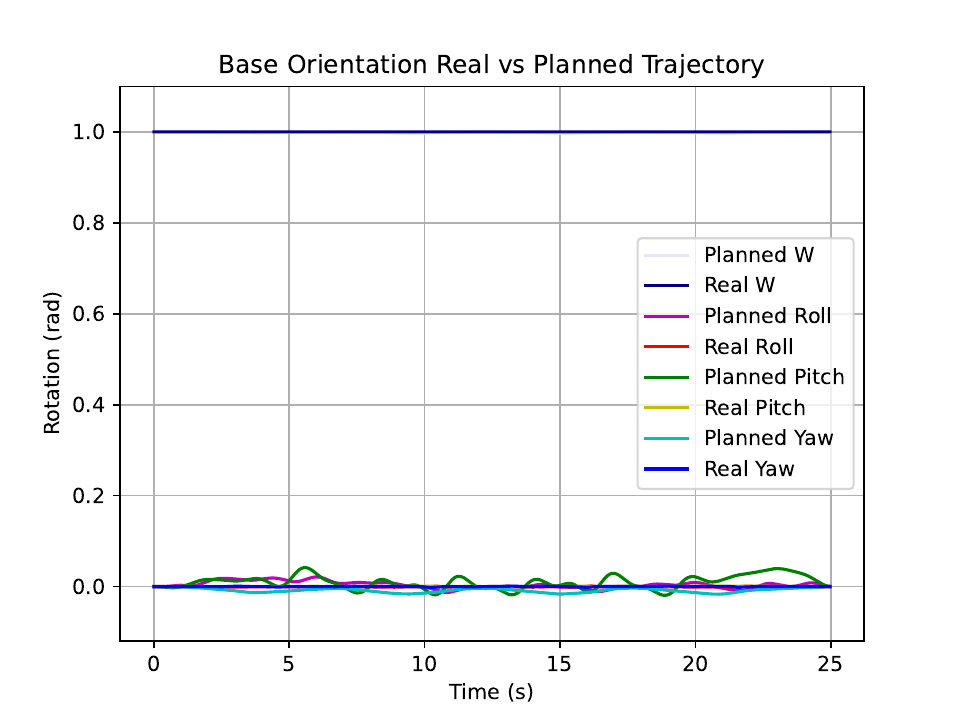}
         \caption{Case 1.2: Body Pose Tracking performance }
         \label{fig:track_body_xy}
    \end{subfigure}    
    \begin{subfigure}{0.67\textwidth}
         \centering
         \includegraphics[width=0.48\textwidth]{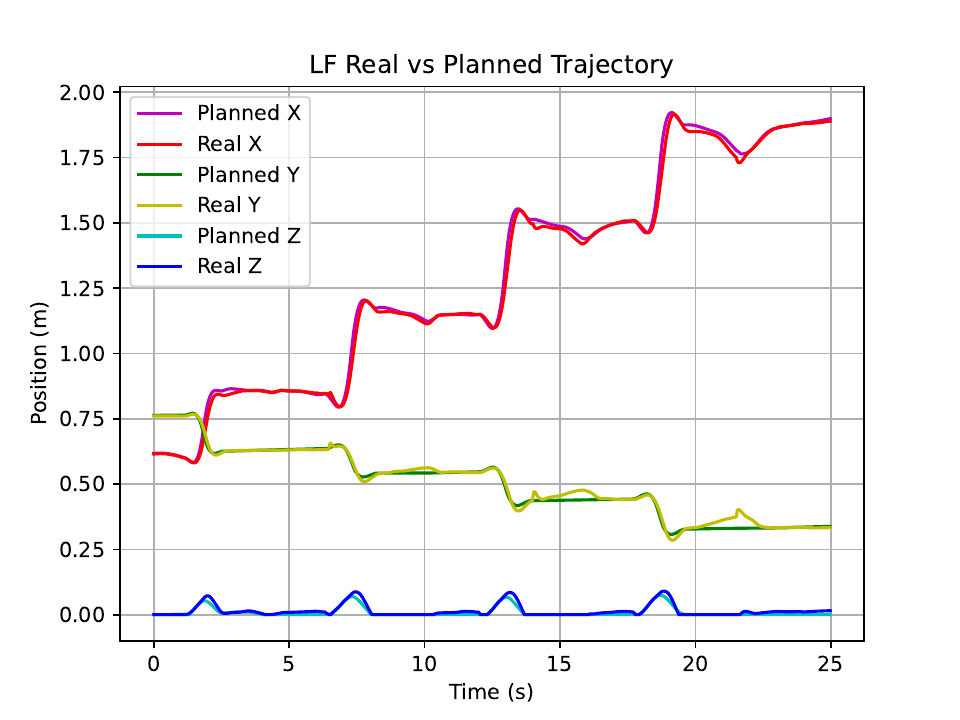}
         \includegraphics[width=0.48\textwidth]{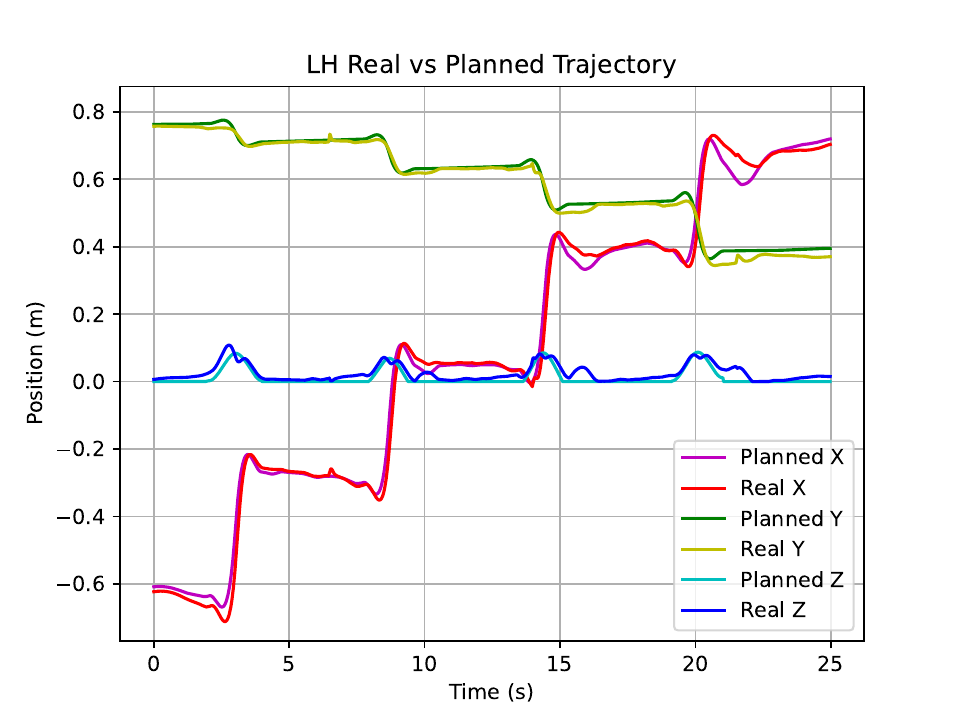}
         \includegraphics[width=0.48\textwidth]{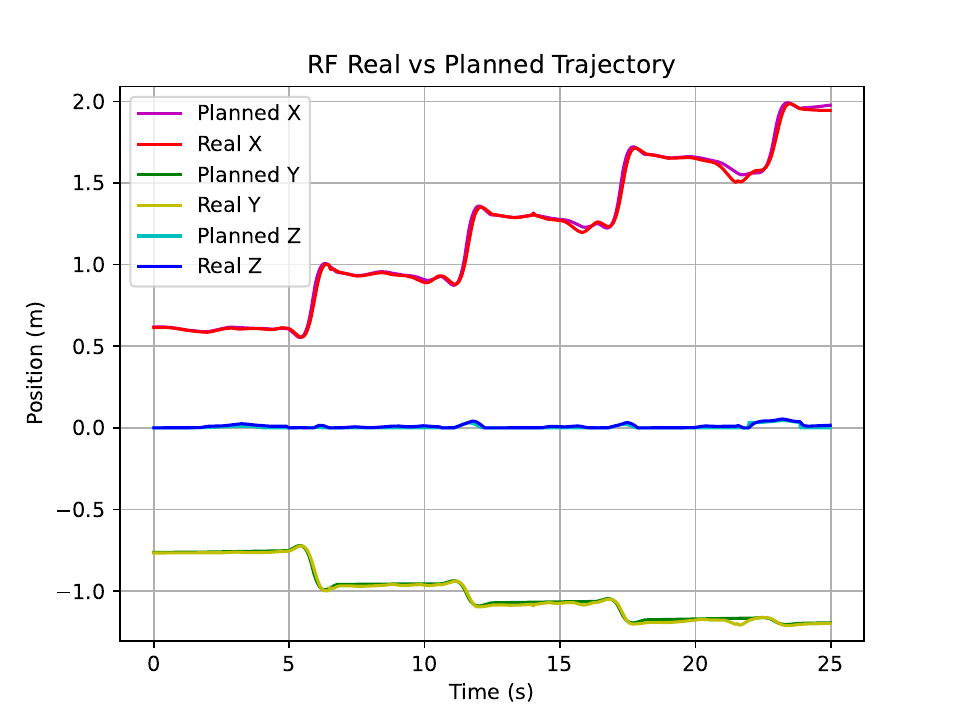}
         \includegraphics[width=0.48\textwidth]{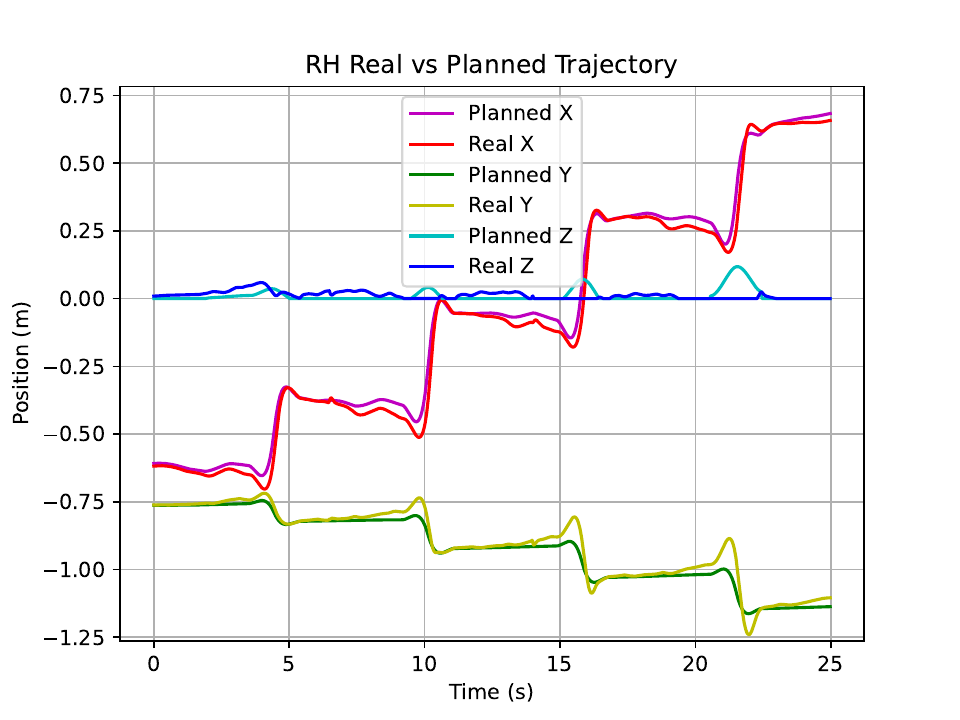}
         \caption{Case 1.2: Arm end-effector tracking performance}
         \label{fig:track_ee_xy}
    \end{subfigure}
\caption{\bf{Case 1.2: Tracking performance using the RL-driven control policy}}
\label{fig:tracking_xy}
\end{figure*}

\cref{fig:tracking_xy} shows the resulting performance of our policy tracking the $XY$ trajectory above. We can see in \cref{fig:track_body_xy} how the body accurately tracks the position displacements in both $x$ and $y$ directions while keeping the $z$ component steady. Regarding arm end-effector tracking, \cref{fig:track_ee_xy} shows how each arm steadily produces motion in both $x$ and $y$ directions, that the policy tracks with small differences despite the significant dynamics mismatch between the model used for planning and that used for the experiments. All in all, we can see that the policy is also able to effectively track a more complex motion profile that combines forward and lateral displacements, accounting for the coupled forces generated by this type of motion.

\subsubsection{Evaluation of number of time-segmented polynomials.} \label{sec:Polys_ev}
\begin{figure}[t]
    \centering
    \includegraphics[width=\columnwidth]{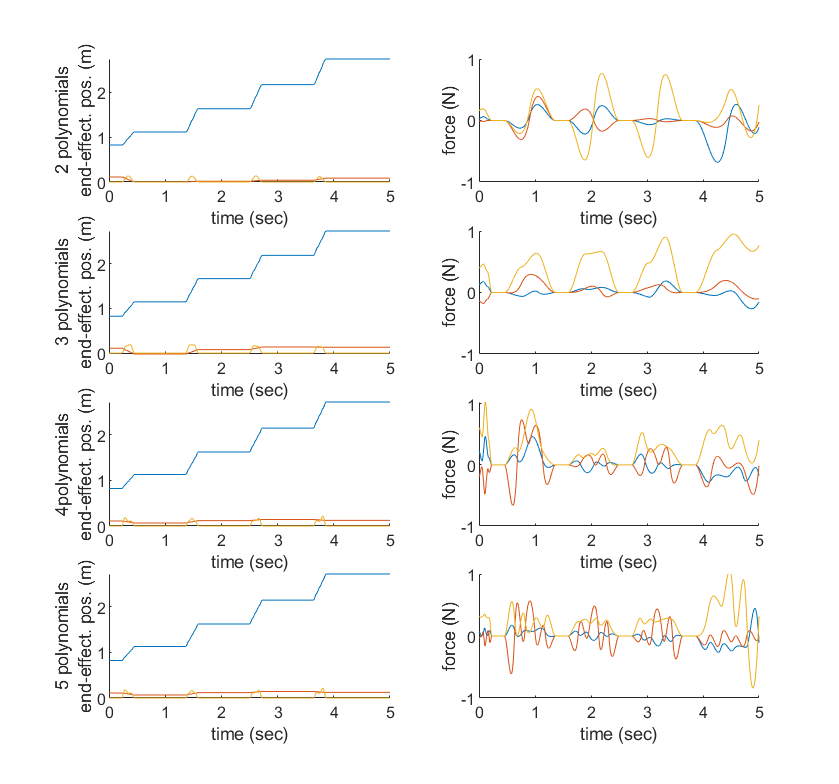}
    \caption{\bf{End-effector position and interaction forces considering 2, 3, 4 a 5 polynomials per segment.}}
    \label{fig:polysperseg}
\end{figure}

\begin{figure}[ht]
    \centering
    \includegraphics[width=\columnwidth]{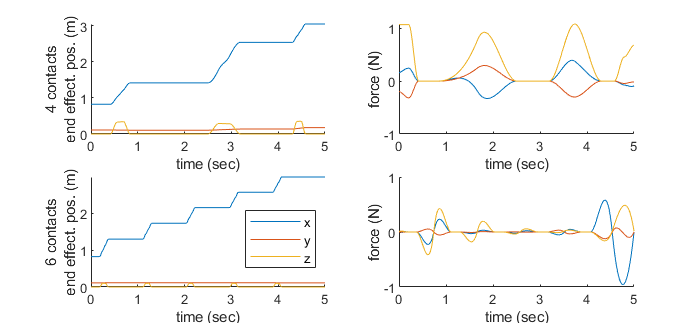}
    \caption{\bf{End-effector position and interaction forces considering 4 and 6 contacts (three polynomials per segment).}}
    \label{fig:contactspertraj}
\end{figure}

As discussed in Section \ref{sec:to}, for a given end-effector trajectory, multiple third-order polynomials are considered during each non-contact phase, and a continuous spline is constructed by combining these polynomials. In addition, for each arm’s force profile, multiple polynomials represent each contact phase. To evaluate how the number of polynomials per segment affects the results, \cref{fig:polysperseg} shows the end-effector position and interaction forces generated by the TO formulation for the first leg when a displacement of 2.5 meters in 5 seconds is required for the robot body. Using three polynomials per segment is considered sufficient to accurately represent both the end-effector positions and forces, achieving good encoding without unnecessarily increasing the number of parameters estimated by the OCP. Furthermore, for the case where three polynomials are used, \cref{fig:contactspertraj} shows the trajectory when 4 and 6 contact phases are considered. In all cases, a valid solution was obtained, reaching the final position; however, greater arm displacements and contact forces are required when the number of phases is reduced. 

\subsection{Case 2: Approximation to the target spacecraft}\label{sec:case2}

\paragraph{Trajectory Optimization:}\label{case2:TO} In this second scenario, the task at hand presents the multi-arm robot free-floating in the space close to the target spacecraft, but not close enough to be able to reach it without the use of the thruster system. We consider this scenario critical, as it mimics situations where the robot may have lost contact with the spacecraft and must recover its position, or possibly a case where the deployment of the robot from a mother spacecraft was not sufficiently precise and the system must correct its position to reach the desired target. 

\begin{figure}[t]
    \centering
    \includegraphics[width=\columnwidth]{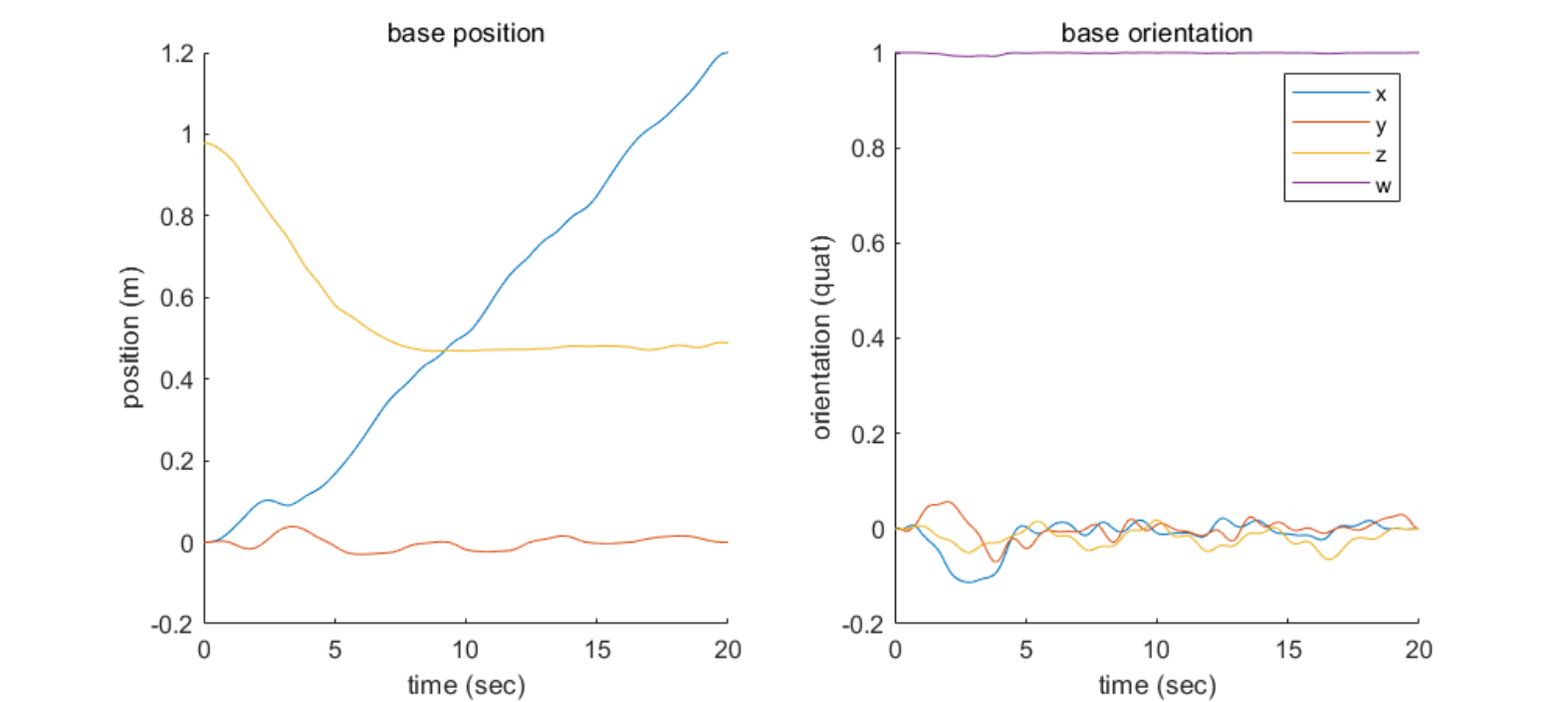}
    \caption{\bf{Position and orientation of the robot base during the trajectory for the second scenario.}}
    \label{fig:case2-bodyposyor}
\end{figure}

The scenario is as described in \cref{sec:setup_description}. The robot must reach the desired target pose in $T=20s$ again, but, this time, starting without contact, with its arms' end-effectors at $0.5m$ height w.r.t the surface. As in the previous case, our proposed TO algorithm computes the target trajectory for the robot body, as well as the motion and force profiles for the end-effectors, together with the thruster forces that must be generated in order to move closer to the surface and then reach the goal pose. \cref{fig:case2-bodyposyor} shows the planned motion for the body. The planner accounts for the need to approach the target surface first, which results in an approximation maneuver at the beginning where the height is reduced in $z$ axis, in order to then proceed as in the previous case and move forward towards the goal.

\begin{figure}[t]
    \centering 
    \includegraphics[width=\columnwidth]{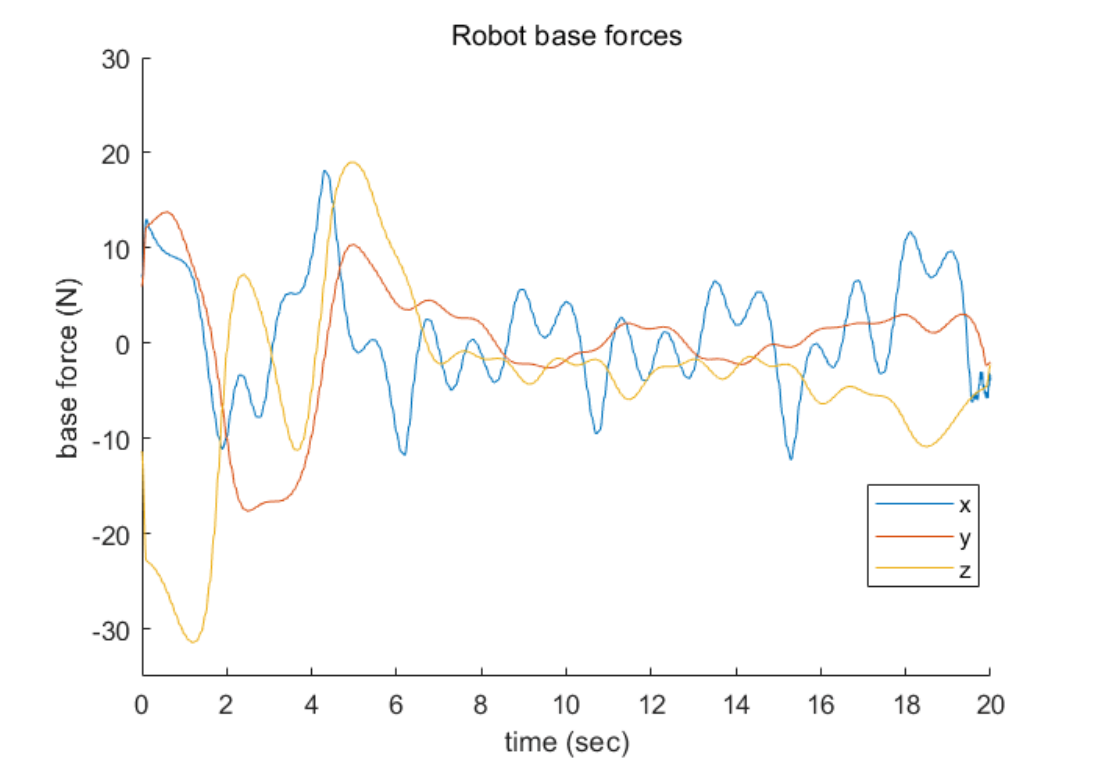}
    \caption{\bf{Robot base thruster forces during the second case trajectory.}} 
    \label{fig:case2-thruster}
\end{figure}

The thruster forces required to execute this maneuver are shown in \cref{fig:case2-thruster}. In this case, the forces are higher in order to approximate the target surface, and then control the overall motion of the robot in order to reach the desired pose. The thrusters make a diagonal approach to the spacecraft during the first 3-5 seconds, and then focus on smoothing the following motion when the arms reach the surface and start taking control of the motion.

\begin{figure*}[ht]
    \centering
    \begin{subfigure}{0.49\textwidth}
         \centering
         \includegraphics[width=\textwidth]{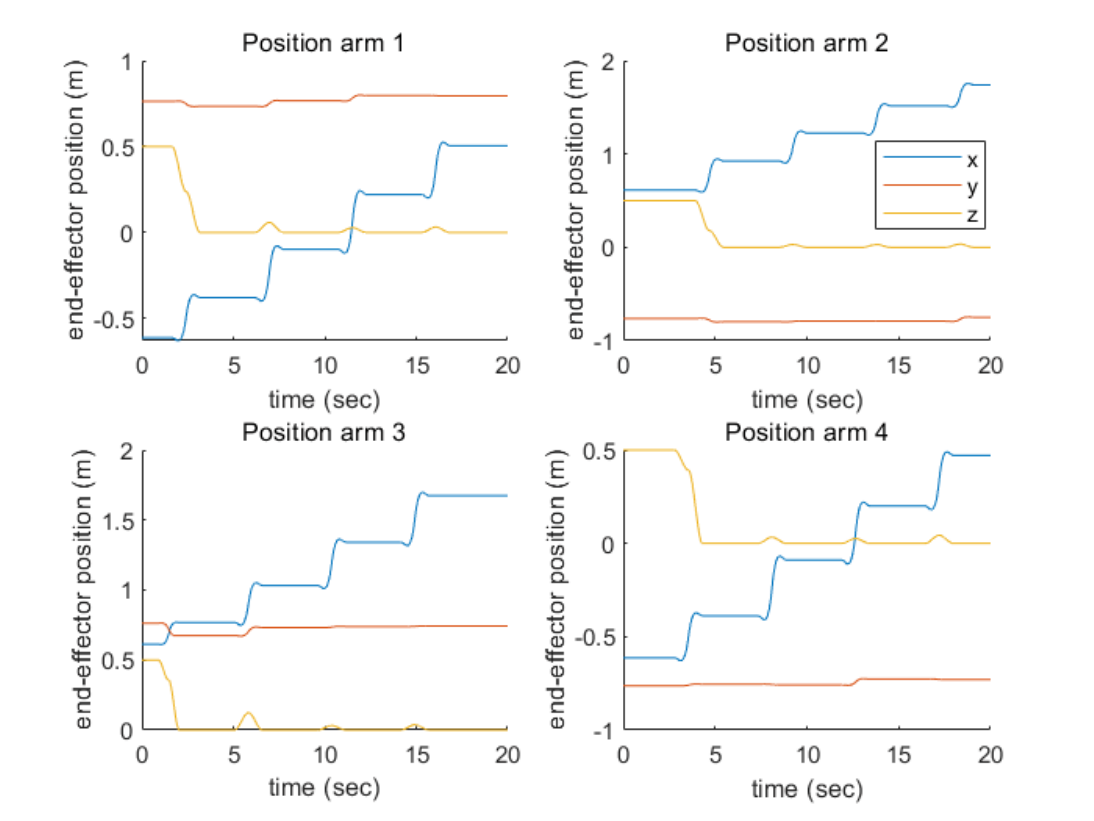}
         \caption{End-effector position during the maneuver in case 2}
         \label{fig:case2-ee_traj}
    \end{subfigure}
    \begin{subfigure}{0.49\textwidth}
         \centering
         \includegraphics[width=\textwidth]{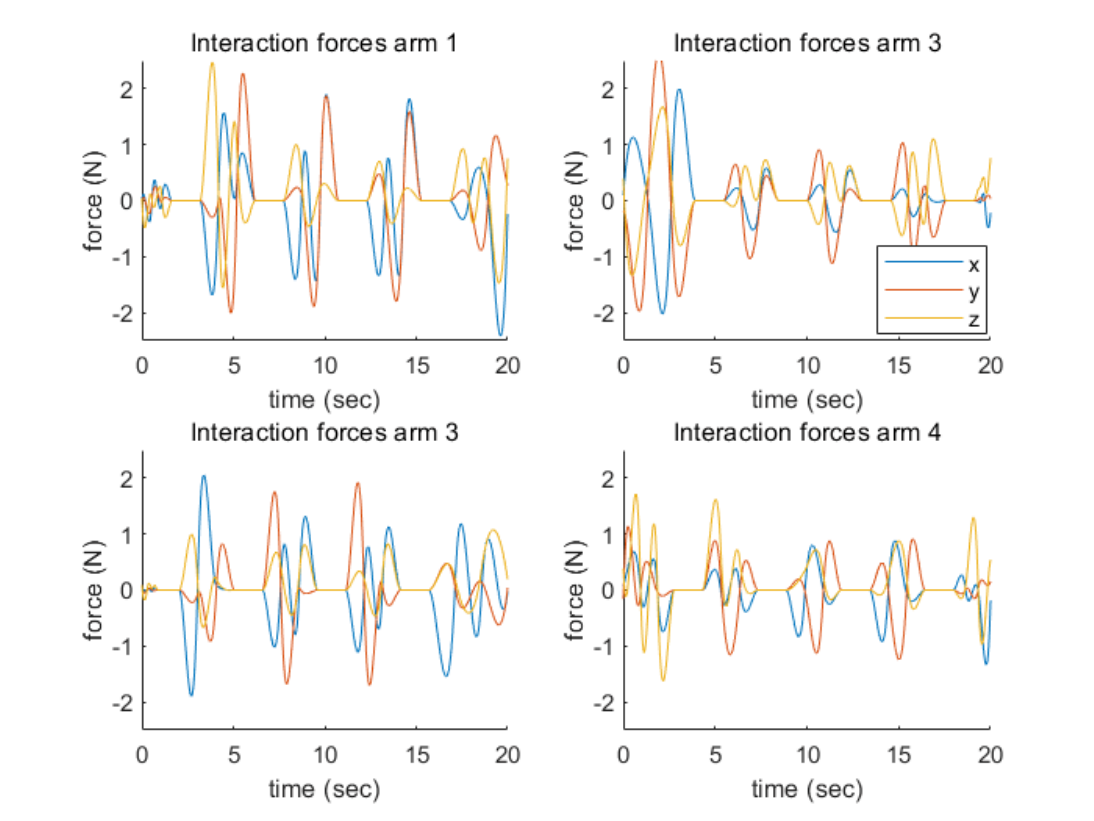}
         \caption{End-effector interaction forces during the second manuever}
         \label{fig:case2-forces}
    \end{subfigure}    
\caption{\bf{End-effector positions and forces during the maneuver}}
\label{fig:case2-ee}
\end{figure*}

Finally, \cref{fig:case2-ee_traj} and \cref{fig:case2-forces} show the planned position and force profiles for the end-effectors of the arms, respectively. Of special interest here is that the position profiles for the $z$ coordinate of every arm start with a first phase of reaching the surface powered by the thrusters, and then switch to a second phase similar to case 1 where the arms take control of the motion and iterate between contact and non-contact phases. 

\begin{figure*}[ht]
    \centering
    \begin{subfigure}{0.33\textwidth}
         \centering
         \includegraphics[width=\textwidth]{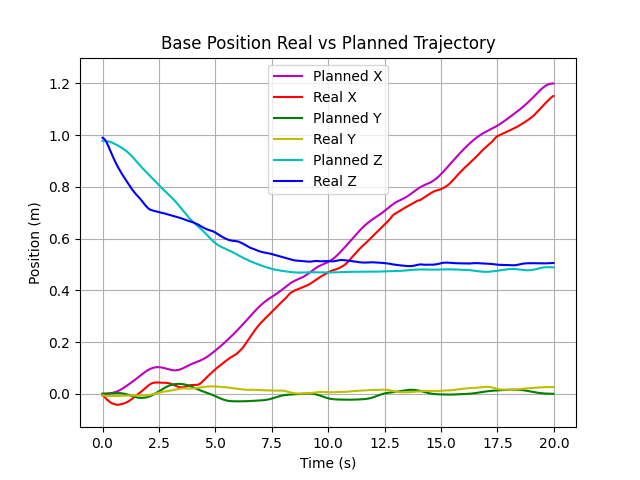}
         \includegraphics[width=\textwidth]{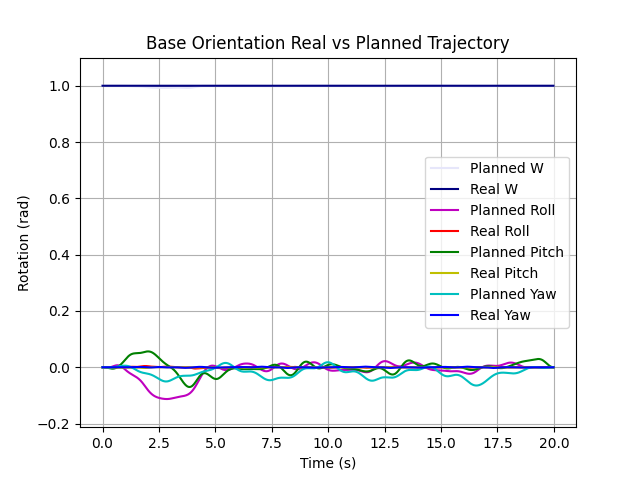}
         \caption{Case 2: Body Pose Tracking performance}
         \label{fig:case2-track_body}
    \end{subfigure}    
    \begin{subfigure}{0.66\textwidth}
         \includegraphics[width=0.5\textwidth]{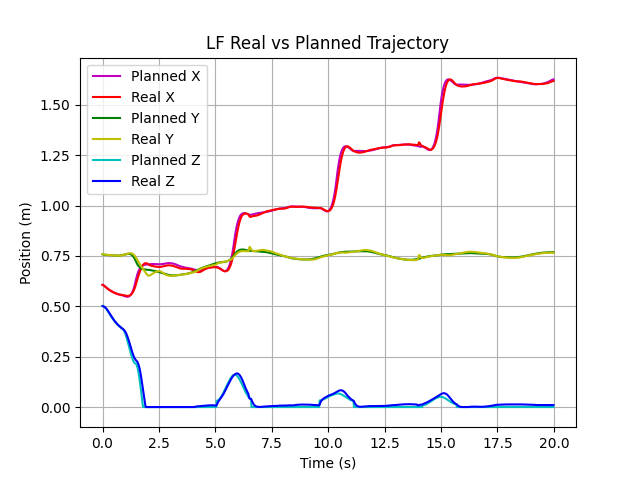}
         \includegraphics[width=0.5\textwidth]{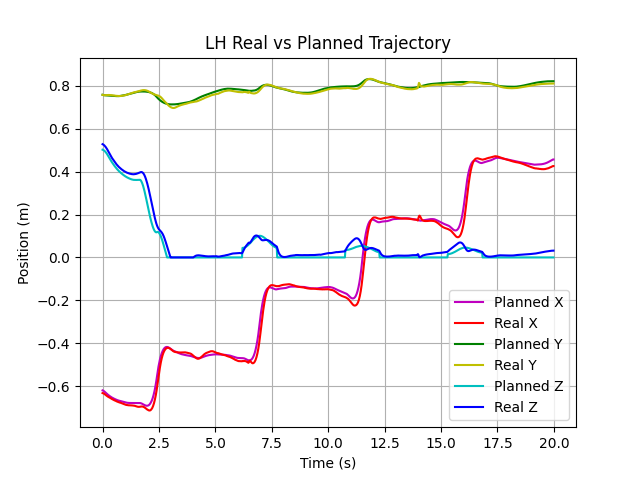}
         \includegraphics[width=0.5\textwidth]{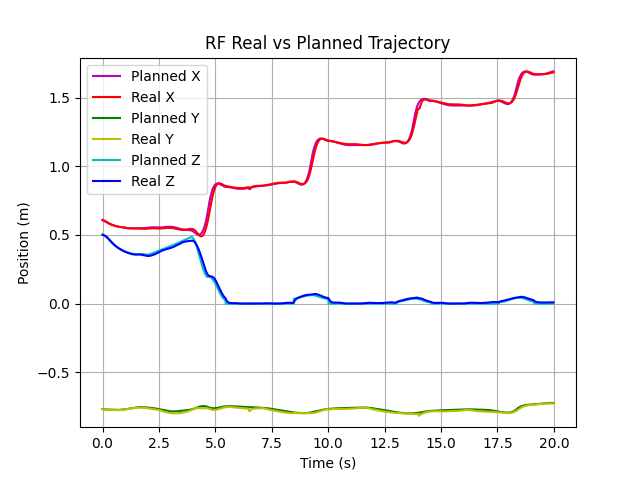}
         \includegraphics[width=0.5\textwidth]{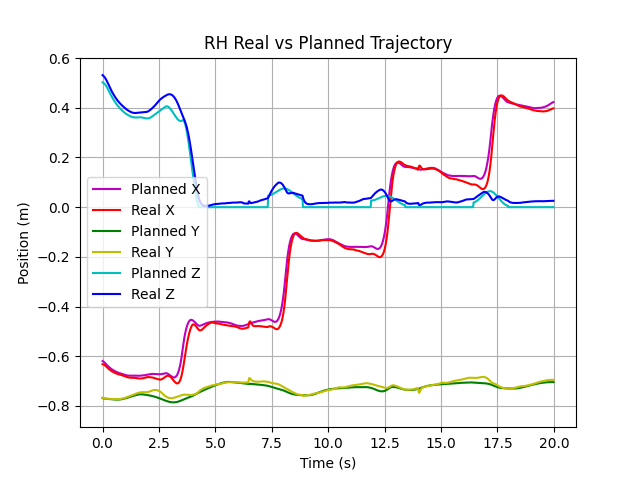}
         \caption{Case 2: Arm end-effector tracking performance}
         \label{fig:case2-track_ee}
    \end{subfigure}
    \caption{\bf{Case 2: Tracking performance using RL-driven control policy}}
    \label{fig:case2-tracking}
\end{figure*}

\paragraph{RL-driven Motion Control:} After illustrating the behavior of our TO approach in the task of approaching the target spacecraft, we now present the results of using the RL-driven controller to follow the motion profile obtained by the planning algorithm. It is worth mentioning that the policy employed here is the same as that used in Case 1, taking advantage of the generic training method described in \cref{sec:setup}. We follow the same subsampling strategy used in case 1, and sequentially feed the policy with the planned targets. \cref{fig:case2-track_body} shows the body pose tracking performance for this second task. The policy is able to track the planned approximation, and then subsequently switch to the arm-based locomotion correctly, with the same delay in tracking the reference explained in the previous case. 

Lastly, we can see the policy's ability to track the desired motion profiles for the arm end-effectors in \cref{fig:case2-track_ee}. The policy successfully tracks the reference motion in both the approximation phase and the subsequent locomotion part, with a particularly good tracking performance in the front arms, and a slightly worse behavior in the hind arms. This difference might be caused by the tendency of the body to move forward, which leaves the hind arms further behind and forces them to take longer steps between docking states, making them prone to higher errors. Despite this, the performance remains high, again demonstrating the RL-driven controller's adaptability to trajectories planned with simplified system models.

\subsection{Tracking performance analysis}

For a more quantitative analysis of the performance of our policy, \cref{tab:rl} summarizes the mean tracking errors during the evaluated trajectories for the base position and each arms' end-effector. The results indicate very low tracking errors, with a few centimeters in the worst case, and of millimeter-scale errors for the best tracking performance, validating the control exerted by the RL policy.

\begin{table}[t]
\small \sf \centering
\caption{\bf Mean tracking errors}
\label{tab:rl}
\begin{tabular}{c|c|c}
\toprule
\bfseries \textbf{Body ID} &  \textbf{Case 1 Error (m)} & \bfseries \textbf{Case 2 Error (m)} \\
\midrule
Base & $1.5 \times 10^{-2}$ & $2.5 \times 10^{-2}$ \\
LF arm & $6.7 \times 10^{-3}$ & $6.2 \times 10^{-3}$ \\
LH arm & $1.1 \times 10^{-2}$ &  $1.3 \times 10^{-2}$ \\
RF arm & $7.3 \times 10^{-3}$ & $7.1 \times 10^{-3}$\\
RH arm & $1.2 \times 10^{-2}$ & $1.5 \times 10^{-2}$ \\
\bottomrule
\end{tabular}
\end{table}

To provide a comprehensive evaluation, we compare our results with those reported in \citep{Belmonte2024SPAICE,RAMON2023102790} for single-arm control, which, for completeness, are also summarized in \cref{tab:compare}. We can see that our policy improves the tracking performance of the arm even when tracking a much more complex motion profile than the controllers compared in previous works. 

\begin{table}[t]
\small \sf \centering
\caption{\bf Control performance comparative}
\label{tab:compare}
\begin{tabular}{c|c}
\toprule
\bfseries \textbf{Controller} & \bfseries \textbf{Mean error (m)} \\
\midrule
PD Controller & $2.7\times{10^{-2}}$ \\
Velocity-based & $2.3\times{10^{-2}}$ \\
Acceleration-based & $1.2\times{10^{-2}}$ \\
Force-based controller & $1.4\times{10^{-2}}$ \\
Optimal control approach & $9.3\times{10^{-3}}$ \\
\textbf{RL-driven (ours)} & $\mathbf{6.7\times{10^{-3}}}$ \\
\bottomrule
\end{tabular}
\end{table}

To further validate the effectiveness of our learning-based control approach, we implemented two classical control methods to evaluate their performance for whole-body control of our four-arm system. First, we implemented a differential inverse kinematics (IK) controller, which enables Cartesian-space control by computing a desired change in joint positions that achieves the desired change in Cartesian pose \citep{buss2004introduction}. This approach suits our Cartesian-space planning algorithm, allowing for seamless substitution of our RL policy, further demonstrating the modularity of our proposed framework. Secondly, we extended the differential IK controller with joint impedance control to assess the performance of our policy, thereby incorporating system dynamics into the computation of the control signal. \citep{KhatibOSC}.

By evaluating these control approaches using the same trajectory generated for Case 1 in \cref{sec:TO_exp}, we report a mean tracking error of $9.2 \times 10^{-2}$ m for the base differential IK controller, and $7.6 \times10^{-2}$ m for the extended version with impedance control. Both strategies result in significantly higher errors than those reported for our learning-based controller in \cref{tab:rl}. In addition, albeit in a different application domain, recent works such as \citep{DAL2025439} report accuracies in the order of $5.1 \times 10^{-2}$ meters for multi-arm robot controllers with similar characteristics to those proposed in this paper. Collectively, these results validate the effectiveness of our approach for high-dimensional control of multi-arm robotic systems in orbital operations.

\subsection{Computational load and suitability for space systems}

The modularity of the presented formulation allows relatively straightforward extension to additional arms by replicating the parameter sets for each new arm while maintaining the overall structure of the optimization problem. Moderate increases in system complexity (e.g., including one or two arms or adding a few joints per arm) result in optimization times that remain practically feasible. Similarly, the RL-based controller scales with the size of the observation and action spaces, which grow with additional arms or joints. While this does not drastically increase computational complexity for moderate scaling, it requires careful tuning of reward weights to balance trade-offs such as power consumption.

We analyze the computational load and real-time feasibility of our approach for space systems separately for the two main components of our framework: the TO planner and the RL tracking policy.

In the current architecture, the TO module serves as an offline planner. It generates a globally feasible reference trajectory (including base motion, arm manipulation, and contact forces) in approximately $10$\ seconds, on a standard workstation (see Sec. 5.1.2). This timeframe is suitable for pre-mission planning or ground-in-the-loop scenarios, where the optimized path is computed beforehand and then uploaded to the robot's onboard controller for execution.

Onboard execution is handled by the RL policy, which is computationally lightweight. The inference step consists solely of matrix-vector multiplications (forward propagation). This low complexity allows the policy to run at high control frequencies ($>100$\,Hz) even on resource-constrained, radiation-hardened processors (e.g., LEON or ARM-based architectures) or to be accelerated via FPGAs. This ensures the system maintains real-time reactivity to orbital perturbations and contact dynamics with minimal power consumption.

%% file: sections/conclusions.tex
\section{Conclusions and Future Work}\label{sec:conclusions}

This paper presented a hybrid strategy for motion planning and control, combining Optimal Control Problem (OCP)-based trajectory planning with reinforcement learning (RL)-driven control, tailored for multi-arm robots in on-orbit servicing applications. We first contextualized our work by analyzing current trends and challenges faced by the space robotics community in motion planning and control for on-orbit operations. Additionally, we reviewed the growing success of RL-based robotic control in terrestrial environments and the initial contributions of RL to planetary-analog environments, and a small number of contributions to the field of orbital robotics, emphasizing the potential advantages of integrating these approaches. Following this, we introduced the architecture of multi-arm robotic systems suited for our framework, while maintaining a generic formulation accommodating any number of arms and DoF per arm, and detailing the kinematic and dynamic formulations essential to the system. A particular focus was placed on the integration of thrusters attached to the robot’s body, which enable direct modulation of the robot’s movement by applying forces to its base.

We then introduced a trajectory optimization strategy that formulates the problem as an OCP. The objective of this algorithm is to generate both the body and end-effector trajectories required to reach a specified target. It also calculates the interaction forces needed at the docking points to facilitate the desired motion and determines the forces to be exerted by the body thrusters to stabilize and enhance the efficiency of the robot’s movement. Next, we described an RL-driven control policy designed to track the trajectory generated by our optimization algorithm. We formulated the Markov Decision Process (MDP) for this task, with particular emphasis on designing an appropriate reward function. After implementing this formulation, we created an RL training environment in which we trained the control policy to track both the robot’s base and end-effector positions. This approach not only enables the policy to follow pre-planned trajectories but also allows it to perform tasks such as reaching specific points with one of the arms. Notably, the policy was trained using a complete system model rather than the simplified centroidal dynamic model employed in the trajectory optimization.

In order to validate the proposed framework, we presented an experimental setup with two different scenarios: One simpler case where the robot is already in contact with the target spacecraft and a second case where the robot needs to leverage the thrusters in order to reach the surface first before attempting to reach the target pose. 

For the first scenario, we evaluated two different trajectories: One with forward motion only, and a second more complex one with combined displacement in $x$ and $y$ directions. For the first task, we generated a sample trajectory with and without the use of thruster forces, and demonstrated the effectiveness of our approach. Our results clearly show that incorporating thrusters significantly improves performance compared to the simpler version of the algorithm, which relies solely on the arms for motion. A key advantage of integrating thrusters is that they significantly reduce the required docking forces from the robot's arms that need to be generated at the docking points, leading to safer and more efficient operation. Additionally, thrusters help to prevent drift during the trajectory and maintain an optimal distance from the target surface. We tested the trained policy on the previously generated trajectory, showing that the robot effectively follows the planned path. This validates the RL policy’s robustness against perturbations and uncertainties, even when tracking a trajectory based on a simplified dynamic model, being able to cope with the inertias and coupled motions resulting from the arms displacements. For the second task with combined direction motion, we also saw that our policy was able to track this kind of trajectories despite the added complexity in terms of arm coordination and additional coupled forces generated by the sideways motion of the arms, further validating the robustness of our RL-driven motion control.

In the second scenario, we repeated the previously described process and showed how our TO algorithm can generate motion profiles that deal with approximation to a target surface starting from a non-contact configuration, and then leverage the arms to reach the desired final pose. The RL policy can track the resulting motion seamlessly as shown in our experiments, thus validating its use in more challenging situations like the one described in this scenario.

In conclusion, this study presents a trajectory optimization framework that advances the state of the art in generating motion paths for free-flying multi-arm robots in on-orbit conditions. Additionally, the RL-driven control policy robustly handles high-dimensional control tasks, despite significant discrepancies between the simplified model of the optimization algorithm and the real system. Future research will aim to further reduce computational complexity in trajectory optimization, facilitating seamless integration into RL training pipelines. Building on this efficiency, a relevant extension is to incorporate a local replanning module operating in a receding-horizon manner on top of the spline-based parametrization. Such a module would adapt the coefficients and durations of upcoming segments in response to environmental changes or state estimation uncertainties, while keeping the global OCP solution as a reference. This would increase adaptability in dynamic environments and improve robustness against unexpected disturbances during on-orbit servicing operations. Investigating the use of thrusters in scenarios such as inter-satellite mobility is a promising avenue for further exploration, as well as solving specific tasks by adding a higher-level task planning stage. Finally, the development of RL techniques with enhanced safety guarantees will be crucial for the continued advance of space robotics applications and will constitute a focus of future research.